%% file: acl_latex.tex
\documentclass[11pt]{article}

\usepackage{acl}

\usepackage{times}
\usepackage{latexsym}
\usepackage{amsmath}
\usepackage{longtable}
\usepackage{pifont}

\usepackage[T1]{fontenc}

\usepackage[utf8]{inputenc}

\usepackage{microtype}

\usepackage{inconsolata}

\usepackage{blindtext}
\usepackage{booktabs}
\usepackage{lscape}
\usepackage{graphicx}
\usepackage{subcaption}
\usepackage{xspace}
\usepackage{bbm}
\usepackage[normalem]{ulem}
\usepackage{multirow}
\usepackage{array,tabularx}
\usepackage{arydshln}
\usepackage{todonotes}
\usepackage{amssymb}
\usepackage[acronym]{glossaries}
\usepackage{makecell}
\usepackage{textcomp}
\usepackage{rotating,booktabs}
\usepackage{cleveref}
\usepackage{subcaption}
\usepackage{enumitem}
\usepackage{soul}
\usepackage{float}
\usepackage{hyperref}
\usepackage{ragged2e}
\usepackage{xcolor}
\usepackage{tcolorbox}
\usepackage{geometry}

\definecolor{teal}{RGB}{0,128,128}
\definecolor{olive}{RGB}{128,128,0}
\definecolor{disgustblue}{RGB}{66,133,244}
\definecolor{highlightgreen}{RGB}{217,234,211}

\title{A Systematic Analysis of Linguistic Features in AI-Generated English Text Detection Across Domains and Models}

\author{Yassir El Attar$^{\blacklozenge1,2}$, Esra Dönmez$^{\blacklozenge1,2}$, Maximilian Maurer$^{\bigstar3,4}$, Agnieszka Falenska$^{\blacklozenge1,2}$ \\
        $^1$Institute for Natural Language Processing, University of Stuttgart\\
        $^2$Interchange Forum for Reflecting on Intelligent Systems, University of Stuttgart\\
        $^3$GESIS - Leibniz Institute for the Social Sciences
        $^4$Heinrich-Heine University Düsseldorf\\
        \textsuperscript{$\blacklozenge$}\texttt{\{yassir.el-attar, esra.doenmez, agnieszka.falenska\}@ims.uni-stuttgart.de} \\
        \textsuperscript{$\bigstar$}\texttt{maximilian.maurer@gesis.org}}

\begin{document}
\maketitle

\input{sections/abstract}

\section{Introduction}
\input{sections/introduction}

\section{Related Work}
\input{sections/related_work}

\section{Methods}
\label{methods}
\input{sections/methods}

\input{sections/results}

\section{Conclusions and Discussion}
\input{sections/conclusions}

\section{Limitations}
Our setup and findings are limited in several ways.
Firstly, while we find particular patterns in English data, this may not hold for other languages. In principle, our methods could be easily applied to any language that (a) has the necessary benchmarks with a broad coverage available, and (b) external tooling to extract the features. The feature extraction tool we use covers a large number of languages, but has some blind spots, particularly regarding low-resource languages and/or languages from the African continent.
Moreover, the applicability of the feature definitions, especially for lexical richness, depends on the particular language. Results in other languages may thus not necessarily be strictly comparable. Extending our analysis to typologically diverse and low-resource languages is therefore an important direction for future work.

Secondly, while we use, to the best of our knowledge, the broadest and most current available benchmark, it does not cover the newest models. Our results thus have limited informativeness on the behavior of these models. The stability of lexical richness in our results suggests it may capture a fundamental property of generated text that persists as models improve, but systematically evaluating this hypothesis on newer frontier models remains an important direction for future work, and underlines the need for continually updated benchmarks to assess whether findings on one set of models still hold.

Thirdly, our experiments cover two important dimensions that impact the detectability of machine-generated text: text domain and model (family). We do not consider prompt variants as a potential source of performance variation. Though outside of the scope of this paper and challenging due to a lack of availability of resources covering all three dimensions, we provide a small pilot experiment including prompt formulation as a factor of performance variation on one of the datasets (CMV) in Appendix~\ref{sec:appendix_cmv_cross_dataset_experiment}, where we find that model-domain alignment matters more than prompt formulation. A systematic study covering complex prompting strategies (personas, stylistic constraints, few-shot examples) is left for future work.

Finally, two further directions extend this work. First, performance variation across text domains -- particularly for conversational text (DialogSum) where removing morphological and information-theoretic features improves detection -- motivates future work on domain-specific feature selection rather than the development of entirely new feature sets. Second, the robustness of lexical richness against adversarial prompting (e.g., instructing models to use diverse vocabulary) is an open question, since classification relies on a combination of features, such interventions may also alter syntactic and semantic properties captured by other feature areas, which we leave for future investigation.

\section{Ethical Considerations}
While our work advances understanding of the features and quality of LLM-generated texts, it raises several important ethical considerations. First, insights from our findings raise potential dual-use concerns. On one hand, they could be misused by malicious actors to develop tools that better mimic human writing patterns. It is therefore essential not only to highlight these risks but also to support community efforts to identify, mitigate, and safeguard against such misuse. The extent to which specific linguistic features can be reliably controlled in generation remains an open question; existing work suggests that fine-grained control over individual linguistic properties is still challenging \citep{alhafni-etal-2024-personalized}. On the other hand, understanding the linguistic signatures of AI-generated text has constructive applications. These insights can help guide generation toward more appropriate content, for example, by suggesting reformulations of user texts to match a desired stylistic profile. Additionally, classifiers based on fully interpretable linguistic features can serve as effective educational tools, helping the public recognize LLM-generated content. Finally, all human-authored texts used in our analyses were drawn from publicly available datasets and handled in accordance with established ethical research standards. No identifiable or private user data was used. Nonetheless, ongoing reflection on issues of consent, data provenance, and user agency remains vital when working with human discourse.

\section*{Acknowledgments}
\label{sec:acknowledgments}
We would like to thank the reviewers for their helpful comments. We acknowledge the support of the Ministerium für Wissenschaft, Forschung und Kunst Baden-Württemberg (MWK, Ministry of Science, Research and the Arts Baden-Württemberg under Az. 33-7533-9 19/54/5) in Künstliche Intelligenz \& Gesellschaft: Reflecting Intelligent Systems for Diversity, Demography and Democracy (IRIS3D) and the support by the Interchange Forum for Reflecting on Intelligent Systems (IRIS) at the University of Stuttgart. We thank the Institute for Natural Language Processing (IMS), University of Stuttgart, for providing the computational resources used in this work.

\bibliography{bib/custom}

\appendix
\section{Methodology}
\input{sections/appendix_A}

\section{Classification Results}
\label{app:classification_results}

\input{sections/appendix_B}

\section{Ablation Results}
\label{app:ablation_results}
\input{sections/appendix_C}

\section{Qualitative Examples}
\label{app:examples}
\input{sections/appendix_qualitative_examples}

\section{Distributional Similarity of Text and Model Domains}
\input{sections/appendix_D}

\section{Use Of AI Assistants}
In this work, GitHub Copilot\footnote{\url{https://github.com/features/copilot}} (version as of November 2025) was used as a code completion/suggestion tool. Additionally, AI-assisted writing tools like Grammarly\footnote{\url{https://app.grammarly.com/}} (accessed November 2025 - January 2026) have been used for spelling checks and grammar corrections.

\end{document}

%% file: sections/abstract.tex
\begin{abstract}
Interpretable linguistic features offer a promising approach for explaining \emph{why} a given text appears machine-generated, particularly for non-expert users. However, existing findings on which features reliably indicate LLM-generated text remain fragmented across feature sets, models, and text domains. To address this gap, we conduct a large-scale empirical study assessing the robustness of linguistic signals for characterizing AI-generated text. Our analysis covers 284 interpretable linguistic features across outputs from 30 LLMs and 10 text domains under cross-model and cross-domain generalization settings. We show that classifiers based solely on linguistic features can reliably distinguish AI-generated from human-written text. However, many previously proposed indicators prove strongly context-dependent, with the exception of measures of lexical richness, which remain robust signals across model families and text domains. These results demonstrate which linguistic signals generalize across contexts and provide a foundation for more reliable, interpretable analyses of AI-generated language.
\end{abstract}

%% file: sections/introduction.tex
Large Language Models (LLMs) are becoming increasingly adept at producing fluent and stylistically \emph{human-like} text. The societal implications of this development are significant: the ease of producing large-scale synthetic text amplifies the risk of misinformation while simultaneously challenging cues used to assess authorship and factuality \citep{srivastava-2025-large}. This creates a growing need for \emph{interpretable explanations} of why a text appears machine-generated, particularly in settings where unreliable or biased detectors can have significant consequences \citep{JIANG2024105070}.

Most existing approaches explain why a text appears machine-generated by interpreting the decisions of black-box classifiers \cite{zellers2019grover, li-etal-2024-mage, guo-etal-2025-profiler}. However, such explanations are often difficult to interpret for non-experts \citep{MILLER20191,kaur2020interpreting} and typically provide only local, example-based insights into the properties of generated text \citep{koike-etal-2026-exagpt}. In response, a growing line of research focuses directly on \emph{interpretable linguistic features}, identifying stylometric and linguistic differences between human and AI-generated writing, such as lexical diversity and syntactic regularity  \citep{munozortiz2024Contrasting, opara2024styloai, reinhart2025DoLLMs}. While these studies provide promising evidence that linguistic features can characterize AI-generated text, the findings remain fragmented across feature sets, LLMs, and text domains. Consequently, it remains unclear which signals reflect general properties of generated language and which arise from particular datasets or model settings \citep{KEHKASHAN2025100793}.

In this work, we take a systematic approach to clarify the fragmented evidence on linguistic signals of AI-generated text. To this end, we first update MAGE \citep{li-etal-2024-mage}, a benchmark for AI-authorship detection across multiple LLMs and text domains, with newer models released since its construction. We then use the updated benchmark to study AI-authorship detection across 30 LLMs and 10 text domains. Our analysis uses 284 interpretable linguistic features, combining and extending features previously studied in stylometry, readability, and AI-text detection. We focus on cross-model and cross-domain generalization, where detectors are tested on unseen models or text domains, and ask three main questions:

\begin{itemize}[leftmargin=0pt,itemsep=0pt,topsep=2pt,label={}]
\item \textbf{RQ1} How robust are linguistic features for distinguishing LLM-generated from human-authored text across models and domains?
\item \textbf{RQ2} Which linguistic features contribute most to distinguishing AI-generated from human-written text, and how does their contribution vary across model families?
\item \textbf{RQ3} How does the contribution of linguistic features vary across text domains?
\end{itemize}

First, we show that classifiers based solely on linguistic features achieve strong performance across models and domains (\S\ref{results:detection}), demonstrating that interpretable linguistic signals can reliably distinguish human- and AI-generated text. Second, feature-area ablations show that \emph{lexical richness} is the most informative feature area overall and the only one whose removal consistently hurts performance on unseen model families, while the contribution of other areas differs across families (\S\ref{results:ablation-domain}). Third, feature-area importance also varies across text domains: lexical richness remains the most consistent signal, whereas the usefulness of other areas depends on the target domain (\S\ref{results:ablation-text-domain}). Taken together, this work provides systematic evidence that linguistic characteristics of AI-generated text exhibit both robust and context-dependent patterns: while some signals, such as lexical richness, generalize across models and domains, others depend on specific generation settings. This highlights the importance of evaluating linguistic explanations under realistic cross-model and cross-domain conditions.\footnote{Our code for the experiments and analyses is available at \href{https://github.com/YassirELATTAR/ling-feats-mgt}{https://github.com/YassirELATTAR/ling-feats-mgt}.}

%% file: sections/related_work.tex
Much of the literature on AI-generated text focuses on the task of \textbf{automatic detection} \cite[][inter alia]{gehrmann-etal-2019-gltr, hao-etal-2025-learning, zhang2024enhancingtextauthenticitynovel, guo2024detectivedetectingaigeneratedtext, ji2024detectingmachinegeneratedtextsjust}, often relying on black-box systems \cite{zellers2019grover, guo-etal-2025-profiler}. Efforts to improve interpretability typically rely on post-hoc explanation methods, such as feature attribution \citep{ribeiro-etal-2016-trust,lundberg-et-al-2017} or example-based explanations \citep{koike-etal-2026-exagpt}. In parallel, several large-scale benchmarks and shared tasks have been proposed to evaluate detection performance across datasets and models \cite[inter alia]{Koike:OUTFOX:2024,li-etal-2024-mage, li-etal-2024-spotting, HSCBZ24, wang-etal-2024-m4gt}. However, the primary focus of this strand of research remains improving detection accuracy rather  than interpretability.

A smaller but growing line of research instead examines \textbf{interpretable linguistic signals} that distinguish AI-generated from human-written language.
For example, \citet{munozortiz2024Contrasting} identify morphosyntactic and semantic indicators that differentiate human- and AI-authored texts in the news domain, while \citet{donmez-etal-2025-ai} analyze LLM-generated counterarguments. \citet{reinhart2025DoLLMs} employ Biber’s 67-feature tagset to explore linguistic patterns in generated text, \citet{opara2024styloai} propose a stylometric detection approach based on handcrafted indicators, and \citet{doughman-etal-2025-exploring} use a small set of linguistic features (POS/NER, readability, and lexical features) for post-hoc analysis of LLM-based detection. These studies demonstrate that interpretable linguistic features can reveal systematic differences between human and AI-generated texts. However, most analyses focus on a limited number of models, domains, or feature sets. Consequently, it remains unclear \textbf{how broadly such linguistic signals generalize across different models and text domains}.

Recent work has also highlighted the importance of \textbf{robustness and generalization} in AI-text detection. Large-scale evaluations show that many detectors and linguistic signals degrade under distribution shift, such as unseen models, domains, or generation settings \citep{li-etal-2024-mage,wang-etal-2024-m4gt,dugan-etal-2024-raid}. Moreover, some apparent linguistic cues may arise from confounding factors such as decoding strategies, prompts, or dataset artifacts rather than properties of machine-generated language \citep{li-etal-2024-mage, reinhart2025DoLLMs}. This raises a broader question: \textbf{to what extent reported linguistic cues reflect genuine properties of AI-generated language}.

%% file: sections/methods.tex
In this work, we train classifiers based on interpretable linguistic features to distinguish between human-written and LLM-generated text. Using these classifiers, we analyze which linguistic signals contribute most strongly to the prediction and evaluate their robustness across models and text domains. In this section, we describe the data  used to train the classifiers, the linguistic features we extract, and the methodological steps taken to analyze linguistic signals of LLM-generated text.

\subsection{Data}
\label{data}

We build on the MAGE benchmark \citep{li-etal-2024-mage} for detecting English LLM-generated text. MAGE contains human-written texts from 10 \textbf{text domains}, covering seven writing tasks: opinion statements (CMV, Yelp), news articles (XSum, TLDR), question answering (ELI5), story generation (WritingPrompts, ROCStories), commonsense reasoning (HellaSwag), knowledge illustration (SQuAD), and scientific writing (SciGen). Corresponding AI-generated texts are produced by 27 models from seven \textbf{model domains} (families): OpenAI, LLaMA, GLM, FLAN-T5, OPT, BigScience, and EleutherAI. MAGE includes generations obtained with three \textbf{prompt types}: continuation, topical, and specified prompts. In addition, it provides a separate \textbf{out-of-distribution (OOD)} test set with human-written texts from four additional domains -- CNN/DailyMail, DialogSum, IMDb, and PubMed --and corresponding texts generated by GPT-4.

For our study, we adapt MAGE to control for prompt variation and to include newer models. First, we restrict the benchmark to generations obtained with \emph{continuation prompts}, in which the model receives the first 30 words of a human-written text and continues it.\footnote{Potential prompt effects are discussed via a cross-dataset
experiment in Appendix~\ref{sec:appendix_cmv_cross_dataset_experiment}} Second, we extend the main benchmark with three models released after MAGE was constructed -- Llama-3.1-8B, Llama-4-Scout-17B-16E, and GLM-5.2 -- bringing the total from 27 to 30 models. The remaining model families are unchanged, as no newer models from these families were available at the time of our experiments. Finally, we extend the OOD portion of the dataset with GPT-5.6-sol generations. A complete list of models and dataset statistics is provided in Appendix~\ref{sec:appendix_dataset_details}.

\subsection{Linguistic Features}
\label{linguistic-features}
We extract \emph{interpretable linguistic features} with the \texttt{elfen} Python package \citep{maurer-2026-elfen} from various groups: surface-level, syntactic, and morphological structures as well as information-theoretic, lexical richness, semantic, and named entity features, and features based on measurements of emotional and psycholinguistic grounding of tokens.
For surface-level, lexical richness, and readability features, we retain only one measure from each set of theoretically equivalent features. For instance, both the type-token ratio (TTR) and the lemma-token ratio capture lexical richness, despite differing in formulation. Appendix~\ref{sec:feature-overview} provides a full overview of the features used in our experiments.

Except for raw counts of tokens,
sentences, characters, lemmas, and syllables, we normalize count-based features (e.g., the number of named entities in a text) by token count to obtain relative frequencies. We remove uniform features (i.e., features with only one value over the dataset) and handle missing and infinite values through feature-specific imputation: binary features are imputed with the mode, integer features with the median, and continuous features with either the mean (for normally distributed features) or median (for skewed distributions), ensuring that the statistical properties of each feature are preserved. The final feature set consists of 284 features: emotion (36), information (2), psycholinguistic (77), dependency (47), lexical richness (3), POS (18), named entities (19), readability (2), surface (6), morphological (58), and semantic (16).\footnote{Feature extraction and preprocessing steps are computed at the document level and fitted on the training data only, ensuring strict separation between train and evaluation splits.}

\subsection{Experiments}

We run two sets of experiments: (1) LLM-authorship detection (\textbf{RQ1}), which evaluates the effectiveness and robustness of classifiers based on interpretable linguistic features, and (2) feature area ablation (\textbf{RQ2} and \textbf{RQ3}), which analyzes the contribution of different linguistic feature groups.

\paragraph{Classifiers}
We use a linear Support Vector Machine (SVM) with balanced class weights to account for the class imbalance between human- and LLM-generated texts in MAGE (details on model choice and parameters in Appendix~\ref{sec:appendix_classifier_details}).

We compare our classifier with three baseline detection methods evaluated by \citet{li-etal-2024-mage} when introducing MAGE: FastText, GLTR, and Longformer. We re-train all three baselines on the same continuation-only extended benchmark used in our experiments to ensure a direct comparison. We omit DetectGPT, which was also evaluated in MAGE, because it requires access to token-level likelihoods of the generating model, which are unavailable for the commercial models in our extended data. Implementation details are provided in Appendix~\ref{app:baselines_reimpl}.

\paragraph{Metrics and Evaluation} We evaluate detection performance using three metrics: \textbf{Macro F1}, which accounts for class imbalance in test sets; \textbf{AUROC} (Area Under the Receiver Operating Characteristic curve) which measures discrimination between classes across decision thresholds; and \textbf{AvgRec} (Average Recall, details in Appendix~\ref{sec:appendix_metrics_details}).

\paragraph{Testbeds}
\label{testbeds}

\begin{table*}[t]
\centering
\scriptsize
\begin{tabular}{@{}p{0.04\linewidth} p{0.27\linewidth} p{0.63\linewidth}@{}}
\toprule
\textbf{Name} & \textbf{Setting} & \textbf{Description} \\
\midrule

\textbf{TB1} & Text domain \& model specific & One classifier per text domain–model pair; trained and tested on data from the same domain–model combination. \\

\textbf{TB2} & Arbitrary text domains \& model domain-specific & One classifier per model domain; trained and tested on all text domains, restricted to AI texts from a single model family. \\

\textbf{TB3} & Fixed text domain \& arbitrary models & One classifier per text domain; trained and tested on all model domains, restricted to texts from a single text domain. \\

\textbf{TB4} & Arbitrary text domains \& arbitrary models & A single classifier trained and tested on all available data across all text and model domains. \\

\textbf{TB5} & Arbitrary text domains \& unseen model domain & Leave-one-model-domain-out evaluation: trained on all data except one model domain and tested on the held-out model domain. \\

\textbf{TB6} & Unseen text domains \& arbitrary models & Leave-one-text-domain-out evaluation: trained on all data except one text domain and tested on the held-out text domain. \\

\textbf{TB7} & Unseen text domains \& unseen models & Trained on all available data and evaluated on new text domains (CNN, DialogSum, IMDb, PubMed) and AI text generated by GPT-4 and GPT-5.6-sol  for the respective domains. \\

\textbf{TB8} & Unseen text domain \& unseen model domain & Leave-one text–model pair out: trained on all data except one text–model domain pair and tested on the held-out pair; similar to TB7 but using existing domains to control for text–model pair effects. \\

\bottomrule
\end{tabular}
\caption{\label{tab:testbeds:desc} Evaluation testbeds used to analyze classifier robustness across text domains and model domains.}
\end{table*}

We evaluate detection performance on eight testbeds from \newcite{li-etal-2024-mage} that vary in generalization scenarios (see summary in \Cref{tab:testbeds:desc} and Appendix~\ref{sec:testbeds_appendix_details}).
While some testbeds operate at fine-grained model granularity (e.g., TB1 evaluates domain–model pairs across all 30 models), we present most analyses at the level of model families to reduce noise from individual model variation and improve interpretability.

\paragraph{LLM-Authorship detection}
Classifiers are evaluated under \textbf{in-domain (ID)} and \textbf{out-of-domain (OOD)} conditions across three evaluation scenarios: general classification, model domain effects, and text domain effects. General classification uses TB4 and TB1 in ID settings, and TB8 and TB7 in OOD settings. Model domain effects are evaluated with TB2 (ID) and TB5 (OOD), while text domain effects are evaluated with TB3 (ID) and TB6 (OOD).

\paragraph{Feature area ablations}
We conduct two types of ablation studies to investigate feature area importance: \textbf{(A)} Leave-one-out, where classifiers are trained on all eleven feature areas but one, and \textbf{(B)} Cumulative,
where feature areas are removed sequentially according to their importance in (A), and classifiers are retrained with progressively fewer feature areas.
We run (A) on all testbeds and (B) only for the domain-agnostic single-classifier settings (TB4 and TB7) to avoid the combinatorial explosion of ablation classifiers.

%% file: sections/results.tex
\begin{table}[t]
\centering
\resizebox{1\linewidth}{!}{%
\begin{tabular}{@{}llrrr@{}}
\toprule
\textbf{Testbed} & \textbf{Method} & \textbf{AvgRec} & \textbf{AUROC} & \textbf{Macro F1} \\
\midrule
\cmidrule{1-5}
\multirow{4}{*}{\textbf{TB4}} & FastText   & .743 & .776 & .741 \\
                              & GLTR       & .556 & .729 & .458 \\
                              & Longformer & .909 & .985 & .907 \\
\cmidrule{2-5}
                              & SVM w/ Ling. Feats. & .838 & .969 & .832 \\
\midrule
\multirow{4}{*}{\textbf{TB7}} & FastText   & .605 & .634 & .605 \\
                              & GLTR       & .569 & .624 & .498 \\
                              & Longformer & .717 & .892 & .703 \\
\cmidrule{2-5}
                              & SVM w/ Ling. Feats. & .812 & .883 & .811 \\
\cmidrule{1-5}
\textbf{TB1} & SVM w/ Ling. Feats. & .954 & .989 & .787 \\
\textbf{TB8} & SVM w/ Ling. Feats. & .808 & .951 & .602 \\
\bottomrule
\end{tabular}
}%

\caption{Classification results: in-domain (TB1 and TB4) and out-of-domain (TB7 and TB8).}
\label{tab:general_results}
\end{table}

\section{LLM-Authorship Detection}
\label{results:detection}

To address \textbf{RQ1}, we first examine general detection performance on TB1, TB4, TB7, and TB8, before using TB2/TB5 and TB3/TB6 to isolate model- and text-domain effects, respectively.

\subsection{General Detection Performance}
\Cref{tab:general_results} presents the classification results. Differences between Macro F1, AUROC, and AvgRec reflect how these metrics are computed. AUROC evaluates score ranking across all thresholds, whereas Macro F1 and AvgRec are computed from predicted labels at the classifier's decision boundary. Under stronger distribution shifts, scores can shift relative to this boundary, so Macro F1 may drop substantially even when ranking performance and balanced recall remain comparatively stable \citep{saito-rehmsmeiter-prec-rec2015,lipton-et-al-optimalf1}.

\paragraph{Aggregate testbeds}
We first consider TB4 and TB7, which provide an overall view of in-domain and out-of-domain detection, respectively. Both testbeds use a single classifier over all relevant text and model domains, making them the most direct settings for comparison with the MAGE baselines.

In the mixed in-domain setting (TB4), where a single classifier is trained and tested on all text and model domains, our linguistic-feature classifier achieves $83.2\%$ Macro F1 and $0.969$ AUROC. It substantially outperforms FastText and GLTR and approaches the performance of Longformer in terms of AUROC ($0.985$). In the fully out-of-domain setting (TB7), which contains unseen text domains and unseen models, our classifier achieves $81.1\%$ Macro F1 and $0.883$ AUROC. Here, it outperforms all three baselines in Macro F1, while its AUROC remains close to that of Longformer ($0.892$). These results show that classifiers based only on interpretable linguistic features can remain competitive with the neural baselines.

\paragraph{Fine-grained testbeds}
We next turn to TB1 and TB8, which provide a more fine-grained view of detection performance across specific text--model combinations. In TB1, classifiers are trained and tested separately for each text domain--model pair. The average performance decreases to $78.7\%$ Macro F1, indicating substantial variation across individual combinations that is hidden by the aggregate TB4 result. TB8 introduces a stronger cross-generalization setting in which both the target text domain and model domain are excluded from training. Performance decreases further to $60.2\%$ Macro F1, showing that some text--model combinations are considerably harder to generalize to than the aggregate OOD setting in TB7 suggests. We examine these model- and text-domain effects in more detail in the next section.

\begin{table}[t]
\centering
\resizebox{.75\linewidth}{!}{
\begin{tabular}{@{}lccr@{}}
\toprule
\textbf{Model Domain} & \textbf{ID} & \textbf{OOD} & \textbf{Difference} \\
       & \textbf{(TB2)} & \textbf{(TB5)} & \textbf{(TB5 - TB2)} \\
\midrule
OPT & .913 & .715 & -.199 \\
Eleuther & .881 & .527 & -.354 \\
LLaMA & .786 & .692 & -.094 \\
FLAN-T5 & .766 & .670 & -.096 \\
BigScience & .656 & .594 & -.063 \\
GLM & .627 & .562 & -.065 \\
OpenAI & .555 & .610 & +.055 \\
\midrule
Mean & .741 & .624 & -.117 \\
\bottomrule
\end{tabular}
}
\caption{Macro F1 results of ID (TB2), OOD (TB5), and their difference per model domain. TB2 trains and tests on the same model domain data with arbitrary domains, while TB5 excludes the target model domain from training.}
\label{tab:testbed2_vs_5_comparison}
\end{table}

\subsection{Model Domain Effects}
We compare model-domain performance in TB2 and TB5 (\Cref{tab:testbed2_vs_5_comparison}). When trained and tested on the data from the same model domain (ID, TB2), the classifiers perform differently for each model domain. For OPT and Eleuther, the classifiers perform above the domain-agnostic ID setting (cf.~TB4 in \Cref{tab:general_results}) by $8.1$ and $4.9$ percentage points, respectively. In contrast, performance for BigScience, GLM, and OpenAI is lower by $17.6$, $20.5$, and $27.7$ points, respectively. This indicates that certain model domains exhibit distinct linguistic signatures that alone provide sufficient signal for the model domain-specific classifiers.

In the OOD setting (TB5), performance drops by $19.9$ and $35.4$ percentage points for OPT and Eleuther, respectively, suggesting that model-domain-specific linguistic patterns in the training data contribute substantially to their detectability. Similarly, for the remaining model domains, we observe a moderate to low performance drop (by $9.6$ to $6.3$ percentage points). Surprisingly, in the OOD setting, the performance increases for the OpenAI models, suggesting that the models display linguistic markers that generalize over LLMs, which can be useful in OOD settings provided sufficient variety in LLM-generated training data.

Finally, we examine whether the model-domain effects observed in TB5 persist when the text domain is also unseen (TB8). \Cref{fig:baseline_testbed8_results_box_plot} (left) shows the distribution of Macro F1 scores across the ten held-out text domains for each model domain. The overall pattern is similar to TB5: performance is highest for OPT and LLaMA and lowest for EleutherAI and GLM. At the same time, the spread within model domains shows that detectability also varies substantially with the accompanying text domain.

\begin{table}[t]
\centering
\resizebox{0.85\linewidth}{!}{

\begin{tabular}{@{}llccr@{}}
\toprule
\multirow{2}{*}{\textbf{Task}} & \textbf{Text} & \textbf{ID} & \textbf{OOD} & \textbf{Difference} \\
       & \textbf{Domain} & \textbf{(TB3)} & \textbf{(TB6)} & \textbf{(TB6 - TB3)} \\
\cmidrule{1-5}
Story&WP & .964 & .878 & -.086 \\
Generation&ROCT & .931 & .698 & -.233 \\\midrule
Commonsense&\multirow{2}{*}{HSwag} & \multirow{2}{*}{.957} & \multirow{2}{*}{.745} & \multirow{2}{*}{-.212} \\
Reasoning&&&&\\\midrule
News&XSum & .945 & .910 & -.036 \\
Articles&TLDR & .916 & .713 & -.203 \\\midrule
Scientific&\multirow{2}{*}{SciGen}& \multirow{2}{*}{.913} & \multirow{2}{*}{.925} & \multirow{2}{*}{+.012} \\
Writing&&&&\\\midrule
Question&\multirow{2}{*}{ELI5} & \multirow{2}{*}{.876} & \multirow{2}{*}{.819} & \multirow{2}{*}{-.057} \\
Answering&&&&\\\midrule
Knowledge&\multirow{2}{*}{SQuAD} & \multirow{2}{*}{.803} & \multirow{2}{*}{.658} & \multirow{2}{*}{-.145} \\
Illustration&&&&\\\midrule
Opinion&CMV & .925 & .875 & -.049 \\
Statements&Yelp & .803 & .783 & -.021 \\
\midrule
\textbf{Mean} &  & .903 & .800 & -.103 \\
\bottomrule
\end{tabular}
}
\caption{Macro F1 results of ID (TB3), OOD (TB6) and their difference. TB3 trains and tests on the same domain with arbitrary models, while TB6 excludes the target domain from training.}
\label{tab:baseline_testbed3_vs_6_comparison}
\end{table}

\subsection{Text Domain Effects}
We observe text-domain effects in TB3 and TB6 (\Cref{tab:baseline_testbed3_vs_6_comparison}). In the ID setup (TB3), variation across text domains is less pronounced than in the model domain experiments, suggesting that text domain effects are not as detrimental to classifier performance as model domain for in-domain classification (cf.~\Cref{tab:testbed2_vs_5_comparison}, ID to \Cref{tab:baseline_testbed3_vs_6_comparison}, ID). For most of the domains, the ID performance is above 90\% F1, and the lowest performance is still relatively high, at 80.3\%.

The OOD experiments reveal several noteworthy patterns: performance in TB6 drops most sharply for ROCT (ROCStories), HellaSwag, and TLDR, by $23.3$, $21.2$, and $20.3$ percentage points, respectively; this is consistent with their larger distributional differences (see Appendix~\ref{sec:wasserstein_text_domains_appendix}).
For SQuAD, the drop is more moderate, at $14.5$ percentage points, while SciGen is the only domain for which performance improves, by $1.2$ points. The remaining domains show smaller performance decreases. Overall, these results indicate that access to text-domain-specific training data can substantially affect detection performance.

\begin{figure}[t]
    \centering
    \includegraphics[width=1\linewidth]{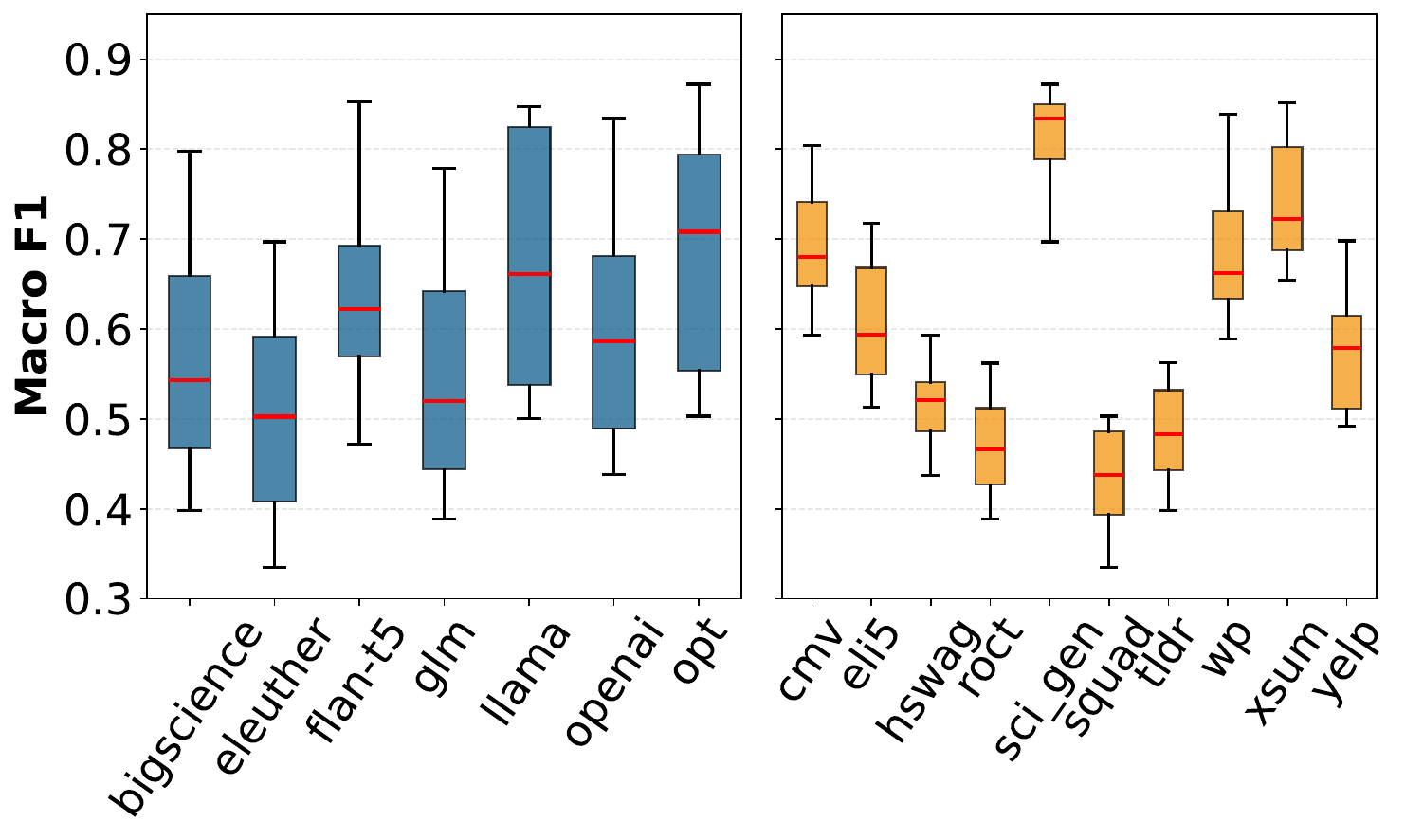}
    \caption{Distribution of Macro F1 scores across \textbf{TB8} (unseen text domain-model domain pairs), left pane for model domains and the right pane for text domains.}
    \label{fig:baseline_testbed8_results_box_plot}
\end{figure}

Finally, we examine text-domain effects when the model domain is also unseen (TB8; \Cref{fig:baseline_testbed8_results_box_plot}, right). Performance varies substantially across text domains, with average Macro F1 ranging from $43.3\%$ for SQuAD to $81.1\%$ for SciGen. This shows that the difficulty of generalizing to an unseen text--model combination depends strongly on the text domain. A similar heterogeneity is visible in TB7, where performance varies considerably across the four novel text domains (see \Cref{tab:baseline_testbed7_table} in Appendix~\ref{app:classification_results}).

\paragraph{Answer to RQ1: Linguistic features provide robust and generalizable signals for distinguishing LLM-generated from human-authored text across models and text domains.} Feature-driven classifiers achieve comparable results to detection methods that rely on neural inference (FastText, GLTR, Longformer) for either extraction or classification, while relying on simple and interpretable representations. Our results indicate the presence of generalizable linguistic markers of LLM-generated text; however, differences in ID and OOD performance highlight substantial influences of model and text domain.

\begin{table*}[t]
\centering
\resizebox{\textwidth}{!}{
\begin{tabular}{@{}llrrrrrrrrrrrrr@{}}
\toprule
  & & \textbf{Baseline} & \textbf{Surface} & \textbf{LexRich.} & \textbf{Emotion} & \textbf{Psycholing.} & \textbf{Read.} & \textbf{Morph.} & \textbf{POS} & \textbf{Depend.} & \textbf{Semantic} & \textbf{Entities} & \textbf{Inform.} \\
\midrule

 \multirow{2}{*}{\textbf{F1 Macro}} &
 \textbf{TB4} & \textbf{.832} & .829 & .689 & .831 & .829 & .833 & .831 & .821 & .830 & .831 & .833 & .817 \\

\cmidrule{2-14}
& \textbf{TB7} & \textbf{.811} & .820 & .575 & .816 & .834 & .810 & .840 & .799 & .824 & .811 & .812 & .828\\
\cmidrule{1-14}
\multicolumn{2}{l}{\textbf{\# Feats. Removed}}  & \textbf{0}  & 6 & 3 & 36 & 77 & 2 & 58 & 18 & 47 & 16 & 19 & 2 \\
\bottomrule
\end{tabular}}
\caption{\label{tab:ablation_testbed4_results_horizontal} Leave-one-out ablation results for TB4 (arbitrary text domains \& arbitrary models) and TB7 (unseen text domains \& model). Classifier performances when each feature area is dropped from the full feature set.}
\end{table*}

\begin{figure*}[t]
    \begin{subfigure}[t]{0.5\linewidth}
        \centering
        \includegraphics[height=1.58in]{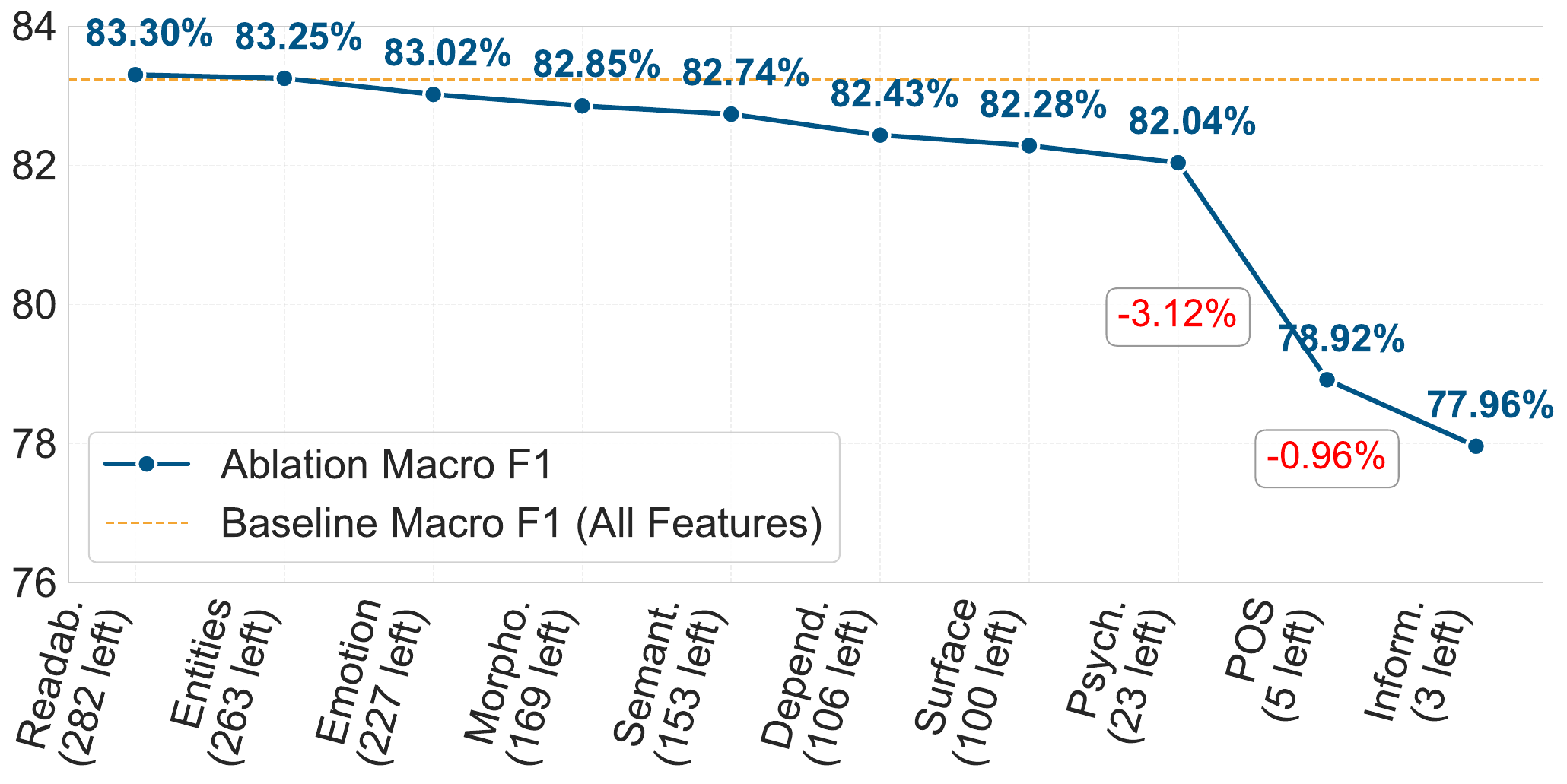}
        \label{fig:ablation_cumulative_testbed4}
    \end{subfigure}%
    \hfill
    \begin{subfigure}[t]{0.5\linewidth}
        \centering
        \includegraphics[height=1.6in]{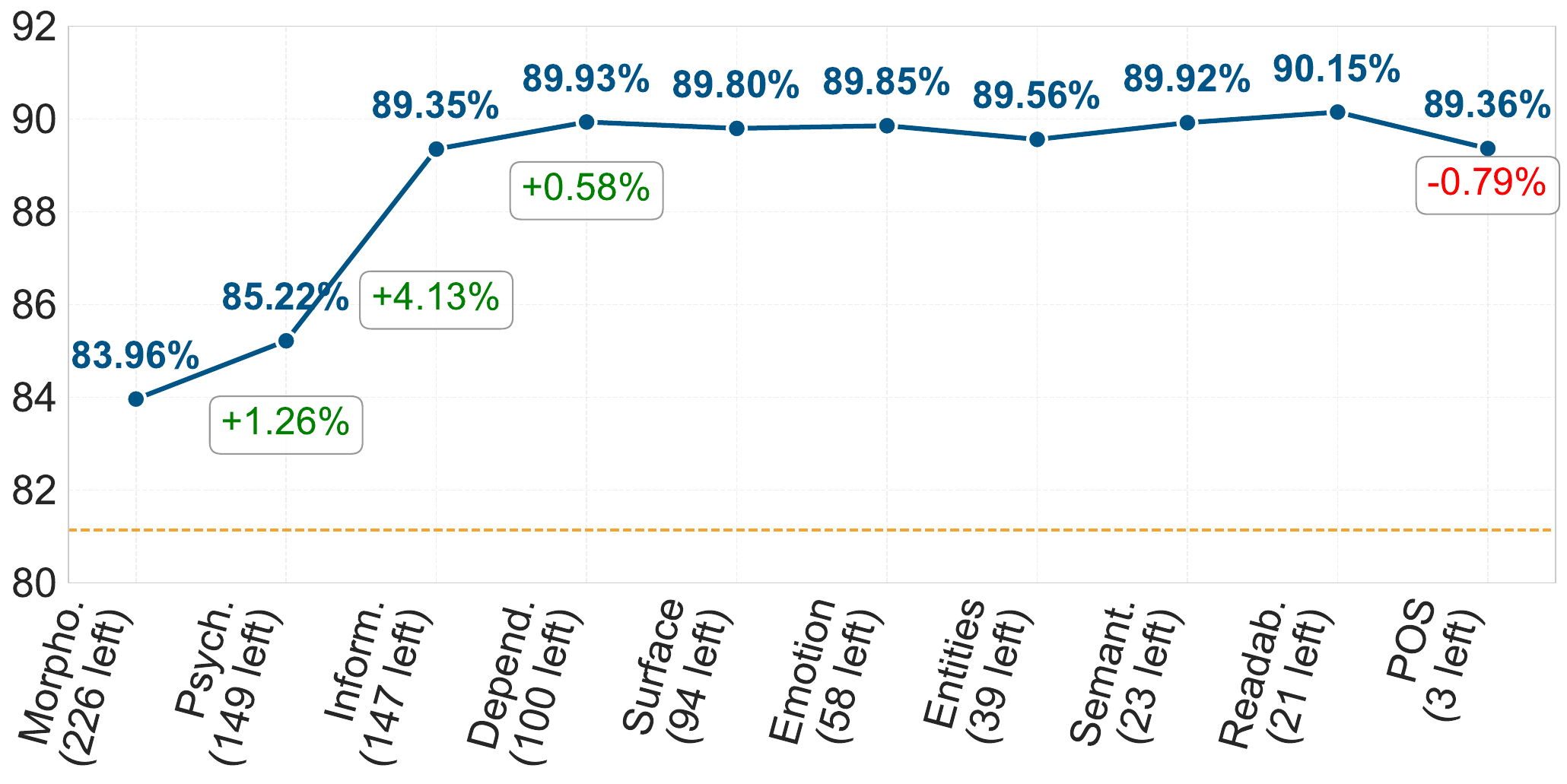}

    \end{subfigure}%
    \caption{\label{fig:ablation_cumulative_testbed7} Macro F1 performance changes with cumulative feature area removal. The left pane shows results for TB4, and the right pane for TB7, in performance impact order from \Cref{tab:ablation_testbed4_results_horizontal}.}
    \label{fig:ablation_cumulative_testbed4_and7}
\end{figure*}

\section{Features of AI-Authored Texts}
\label{results:ablation}

We now zoom in on the effects of the types of linguistic features on the classification performance. We start by discussing the general classifiers and then focus on the model (\textbf{RQ2}) and text domains (\textbf{RQ3}).

\subsection{Overall Feature-Area Importance}

\Cref{tab:ablation_testbed4_results_horizontal} displays the results from the leave-one-out feature-area ablation on the ID domain-agnostic classifier (TB4) and the OOD setting (TB7).

\paragraph{ID feature importance}
In the TB4 setting, we make two observations. First, dropping \emph{lexical richness} features results in a stark performance drop of $14.3$ percentage points, far larger than for any other feature area. The next two most influential areas are \emph{information} and \emph{POS}, with $1.5$ and $1.1$ drops, respectively. Second, the effects from the remaining eight feature areas are marginal, with the largest decrease of $0.3$pp. To examine how performance changes as multiple feature areas are removed, we additionally conduct a cumulative ablation, removing feature areas sequentially according to their importance in the leave-one-out analysis. The resulting pattern is consistent with the leave-one-out results (left pane of \Cref{fig:ablation_cumulative_testbed4_and7}). Dropping the \emph{readability} area initially improves performance, while removing \emph{entities} leaves it essentially unchanged. As the remaining areas are removed one by one, the performance slowly degrades.
However, even after removing ten areas, the classifier retains much of its performance, dropping only about $5$ percentage points from the full-feature baseline (from $83.2\%$ to $77.96\%$).

\paragraph{OOD feature importance}
In the OOD setting (TB7 in \Cref{tab:ablation_testbed4_results_horizontal}), the effect of removing \emph{lexical richness} is similarly pronounced. Among the remaining feature areas, removing \emph{POS} features decreases performance by $1.2$ percentage points, while removing \emph{readability} or \emph{semantic} features has little effect. Removing several other feature areas improves performance, with the largest gain observed for \emph{morphological} features ($+2.9$ points).

Unlike the ID setting, the cumulative ablation in TB7 shows that removing several feature areas can improve OOD performance (right pane of \Cref{fig:ablation_cumulative_testbed4_and7}). Removing the first four areas increases Macro F1 from $81.1\%$ to $89.9\%$ ($+8.8$ percentage points), after which performance remains relatively stable as further areas are removed. Notably, the classifier using only \emph{lexical richness} features still achieves $89.36\%$ F1, substantially above the full-feature baseline. This suggests that several feature areas capture signals that do not generalize well to the unseen domains, whereas lexical richness provides a particularly robust OOD signal.

\begin{figure}[t]
    \includegraphics[width=\linewidth]{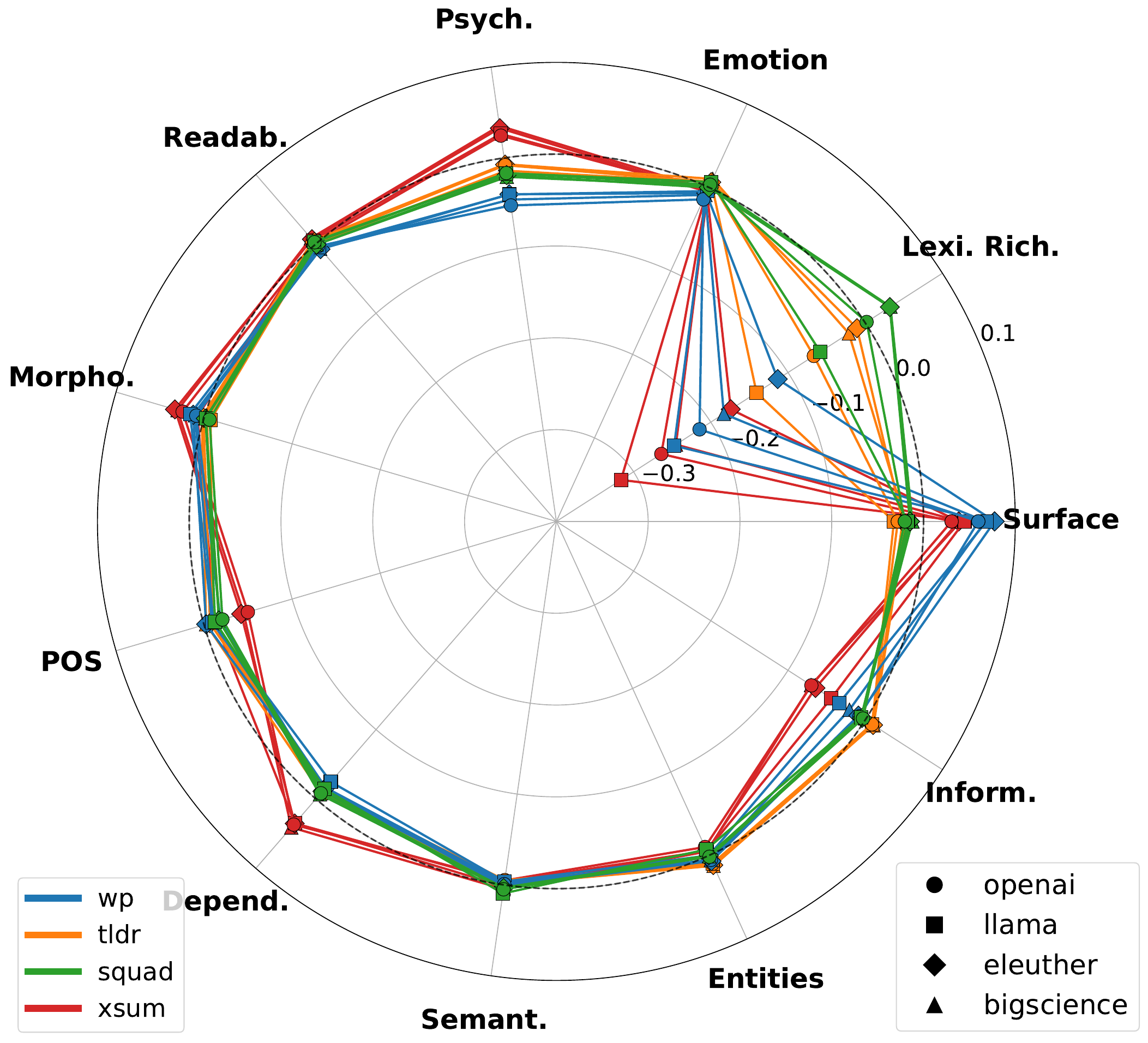}
    \caption{Ablation performance differences across 16 text domain-model domain after removing the corresponding feature area (selection based on Appendix~\ref{sec:testebed8_selection_16_pairs}).}
    \label{fig:spiderweb_plot_ablation_for_16_domain_family_pairs}
\end{figure}

\paragraph{Joint text--model shift}
We next examine whether the observed patterns hold across individual unseen text--model pairs in TB8.
A complete leave-one-out ablation over all 70 text-domain--model-domain pairs would require 770 classifiers. We therefore select four text domains and four model domains based on the performance analysis (see Appendix~\ref{sec:testebed8_selection_16_pairs} for details), yielding 16 representative pairs. \Cref{fig:spiderweb_plot_ablation_for_16_domain_family_pairs} shows two main patterns. First, removing \emph{lexical richness} leads to the largest performance drops for most pairs, although its effect is much weaker for SQuAD. Second, the effects of the remaining feature areas are smaller (or even reversed) and vary across text--model pairs, with their removal sometimes improving and sometimes reducing performance. This indicates that, beyond lexical richness, the usefulness of linguistic features depends strongly on the particular text and model domain. Detailed pair-level results are provided in Appendix~\ref{sec:lexical_richness_results_testbed8}.

\begin{figure*}[t]
    \centering
    \includegraphics[width=.96\linewidth]{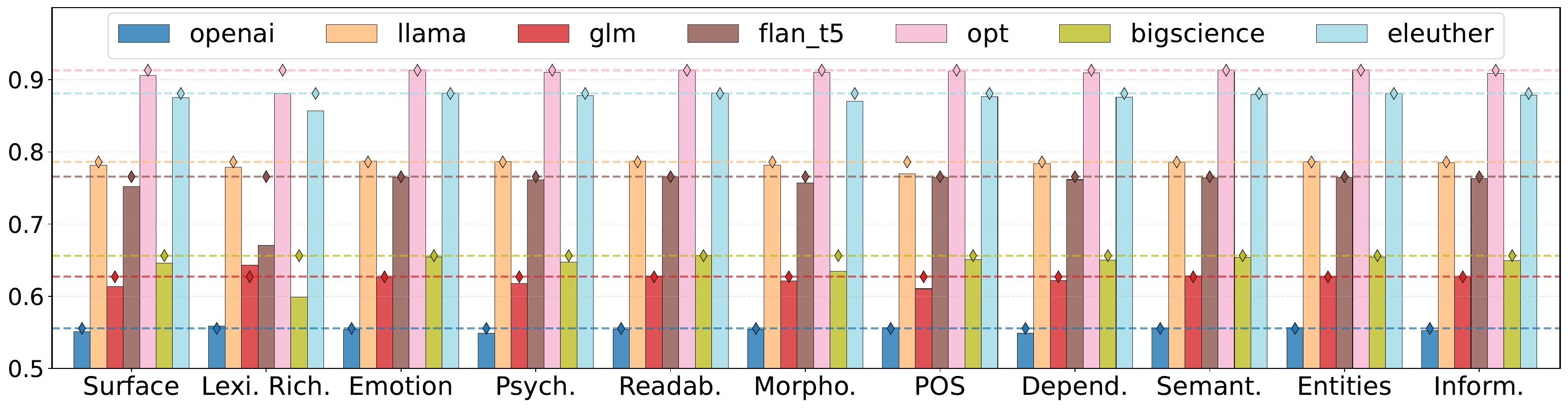}
    \caption{Ablation results on TB2. The horizontal dashed lines indicate the baseline results (all features) and the bars indicate the performance in Macro F1 after removing the corresponding feature area.}
    \label{fig:ablation_results_testbed2_figure2}
 \end{figure*}

\begin{figure}[t]

    \centering
    \includegraphics[width=1\linewidth]{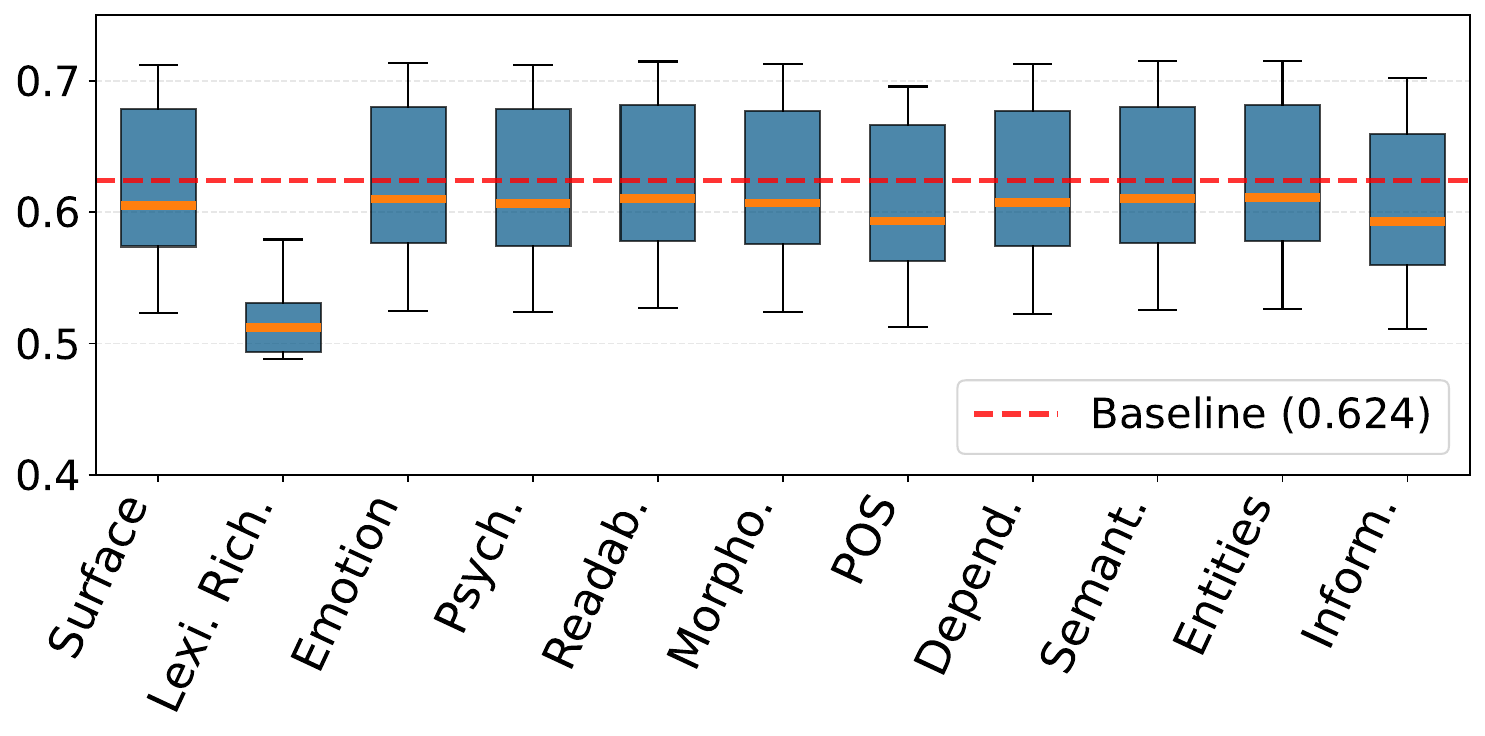}
    \caption{Ablation performance ranges on TB5 across the 11 feature areas after removing the corresponding feature area compared to the averaged baseline.}
    \label{fig:ablation_results_testbed5_box_plot}
\end{figure}

\subsection{Feature-Area Importance Across Model Domains}
\label{results:ablation-domain}

\Cref{fig:ablation_results_testbed2_figure2} shows the feature-area ablations for the ID model-domain setting (TB2). The effect of \emph{lexical richness} varies substantially across model domains: removing it slightly improves Macro F1 for OpenAI ($+0.3$ points) and GLM ($+1.6$), but reduces performance for the remaining families, with the largest drops for FLAN-T5 ($-9.5$) and BigScience ($-5.7$). Removing most other feature areas has comparatively small effects. This suggests that the linguistic signals useful for in-domain detection differ across model families.

The pattern is more consistent in the OOD setting (TB5; \Cref{fig:ablation_results_testbed5_box_plot}). Removing \emph{lexical richness} reduces performance for every held-out model domain, by between $3$ and $18$ percentage points, with the largest drop for LLaMA. By comparison, removing \emph{information} features decreases performance by only $1.2$--$2.4$ points. This indicates that lexical richness provides a particularly robust signal for generalization across model domains (complete ablation results are reported in \Cref{tab:ablation} in Appendix~\ref{app:ablation_results}). To provide qualitative intuition for these patterns, Appendix~\ref{app:examples}  presents representative human- and LLM-generated examples illustrating differences in the selected feature areas.

\paragraph{Answer to RQ2:
\emph{Lexical richness} has the largest overall impact on classifier performance, emerging as the strongest indicator of LLM-generated text.} Feature ablations across model domains, however, reveal substantial variation between model families: the contribution of individual feature areas differs across families and between ID and OOD settings. In particular, while the importance of \emph{lexical richness} varies across model domains in the ID setting, it provides a consistently useful signal when generalizing to unseen model domains. We further examine these model-family differences through an analysis of lexical-richness distributions in Appendix~\ref{sec:feats_distribution_analysis}.

\begin{figure*}[t]

    \begin{subfigure}[t]{\linewidth}
        \centering
        \includegraphics[width=\linewidth, trim=0 0 0 0, clip]{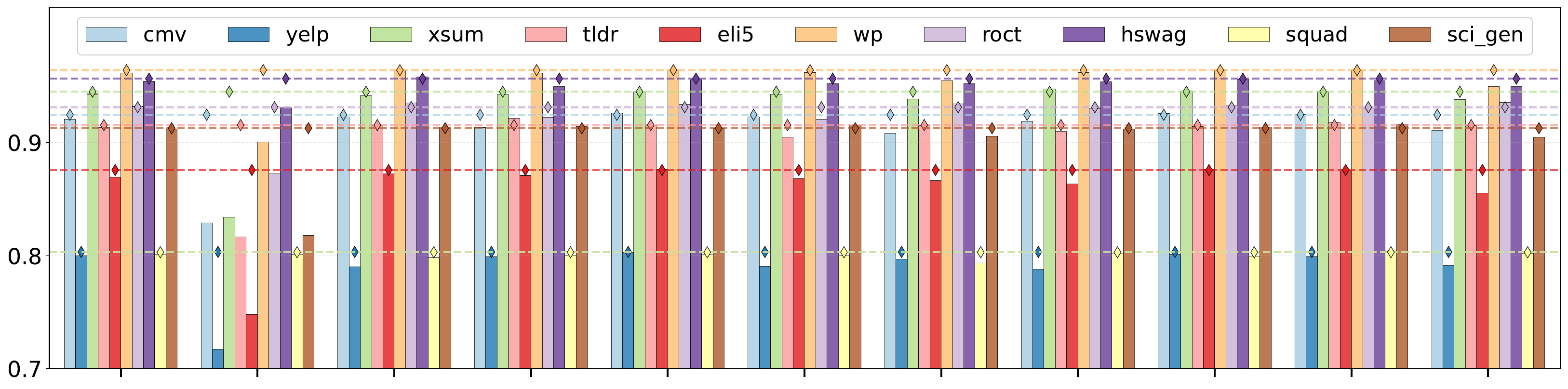}
         \label{fig:ablation_results_testbed3_figure}
    \end{subfigure}%
    \vspace{-0.4cm}
    \begin{subfigure}[t]{\linewidth}
        \centering
        \includegraphics[width=\linewidth, trim=0 0 0 0, clip]{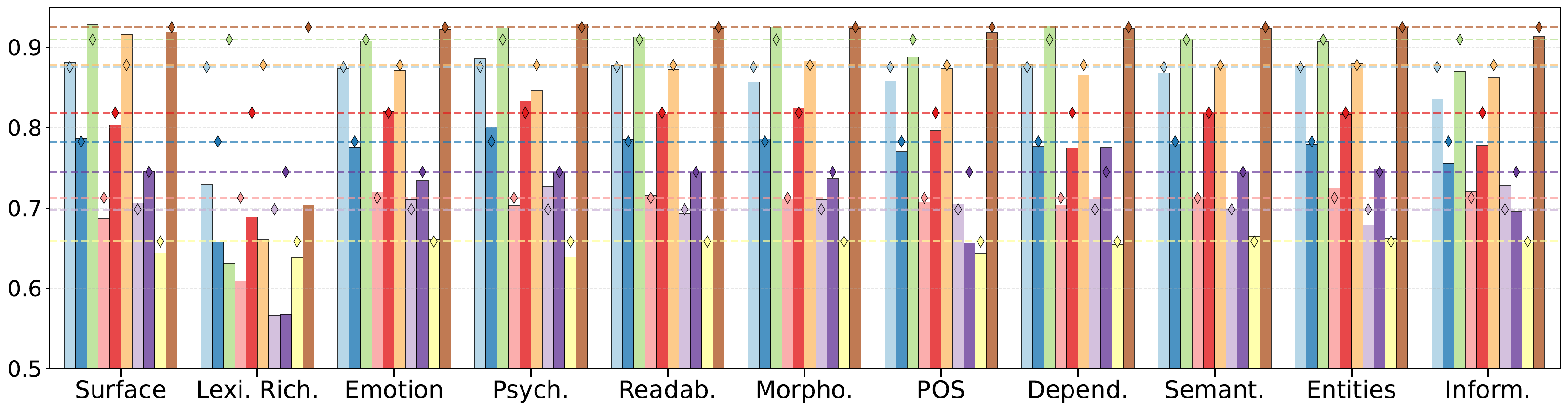}
        \label{fig:ablation_results_testbed6_figure}
    \end{subfigure}

    \caption{Macro F1 results of ablation study, top pane: TB3 and bottom pane: TB6. The horizontal dashed lines indicate baseline results for each model, and the bars the performance after dropping the corresponding feature area.}
    \label{fig:ablation_results_testbed3_and_6}
\end{figure*}

\subsection{Feature-Area Importance Across Text Domains}
\label{results:ablation-text-domain}
\Cref{fig:ablation_results_testbed3_and_6} shows the feature-area ablation results for the ID and OOD text-domain settings. In the ID setting (TB3; upper pane), \emph{lexical richness} has the largest and most consistent effect: removing it decreases Macro F1 across all ten text domains. The effect is negligible for SQuAD (-0.2 percentage points) and relatively small for HellaSwag (-2.6), but reaches 12.8 points for ELI5. For most other feature areas, the effects are smaller and vary across text domains, indicating that their usefulness is more domain-specific.

This pattern becomes stronger in the OOD setting (TB6; lower pane). Removing \emph{lexical richness} decreases performance across all ten held-out text domains, with drops ranging from $2.0$ percentage points for SQuAD to $27.8$ points for XSum. The effects of the remaining feature areas are again mixed and depend on the target domain. For example, removing \emph{surface} features improves performance for WritingPrompts and XSum, but decreases it for TLDR and ELI5. Overall, these results suggest that \emph{lexical richness} provides the most consistent signal across text domains, whereas the contribution of other linguistic feature areas is more domain-dependent.

\paragraph{Answer to RQ3: Feature-area importance varies across text domains.}
\emph{Lexical richness} is the most consistently influential feature area, while the importance of the remaining areas varies substantially across text domains (further analyses on this feature area in Appendix~\ref{sec:lexical_richness_results_testbed8}). To further examine these differences, we compare feature distributions across text domains using Wasserstein distances (Appendix~\ref{sec:wasserstein_text_domains_appendix}). The resulting distributional differences provide additional context for the variation in feature informativeness and cross-domain detection performance.

%% file: sections/conclusions.tex
LLM-generated text increasingly populates both formal and informal discourse. Understanding the linguistic markers that signal such text is therefore essential, both to develop effective detection methods and to provide interpretable explanations to participants of such discourse.

Through extensive classification experiments, we showed that linguistic features (interpretable to both experts and non-experts) are highly informative of whether a text is generated by an LLM. At the same time, the linguistic markers of generated text are far from uniform. Many indicators previously proposed in the literature turn out to depend strongly on particular models or text domains. In contrast, measures of lexical richness consistently emerge as robust signals across contexts. Beyond individual features, systematic patterns also appear across groups of LLMs and types of text domains, suggesting that both model family and discourse context shape the characteristics of generated language.

These findings highlight both the promise and the limitations of linguistically grounded explanations of AI-generated text. While some signals generalize across contexts, others depend on specific models or domains. As a result, detection systems deployed in real-world settings, such as recognizing LLM-generated peer reviews or papers, must be evaluated under diverse conditions and updated as new models emerge. More broadly, our results emphasize the importance of systematically assessing the robustness of linguistic signals when using them to interpret or detect AI-generated language.

%% file: sections/appendix_A.tex
\subsection{Dataset Details}
\label{sec:appendix_dataset_details}

Additional dataset statistics for the MAGE data used in our experiments are reported in Table \Cref{tab:mage_data_statistics}. The 30 models in the main benchmark belong to seven model families: OpenAI (GPT-3.5-turbo, text-davinci-002/003; parameter counts undisclosed), LLaMA (7B, 13B, 30B, 65B, Llama-3.1-8B, and Llama-4-Scout-17B-16E, with 17B activated and 109B total parameters), GLM (GLM-130B and GLM-5.2, with approximately 753B total parameters), FLAN-T5 (small: 77M, base: 248M, large: 0.8B, XL: 3B, XXL: 11B), OPT (125M, 350M, 1.3B, 2.7B, 6.7B, 13B, and 30B, including 1.3B and 30B IML variants), BigScience (BLOOM-7B1: 7.1B, T0-3B: 3B, T0-11B: 11B), and EleutherAI (GPT-J-6B: 6B, GPT-NeoX-20B: 20B).

For the newly added models, we follow the MAGE generation procedure \citep{li-etal-2024-mage}: each model receives the first 30 words of a human-written text and generates a continuation. We use the same 10 text domains and preserve the original train/validation/test split assignments. Three generations for ROCStories were identical to their human-written source texts and were discarded.

For the OOD portion of the dataset, we additionally generate texts with GPT-5.6-sol for the same four domains and human-written instances used in the original GPT-4 set: CNN/DailyMail, DialogSum, IMDb, and PubMed. This allows both models to be compared on the same human-written texts.

\begin{table*}[t]
\centering
\resizebox{0.9\textwidth}{!}{%
\begin{tabular}{lcccccr}
\toprule
\textbf{Split} & \textbf{CMV} & \textbf{Yelp} & \textbf{XSum} & \textbf{TLDR} & \textbf{ELI5}\\
\midrule
Train & 4,223H / 22,788AI & 31,827H / 22,929AI & 4,708H / 28,451AI & 2,826H / 22,211AI & 16,706H / 27,948AI \\
Val & 2,436H / 2,797AI & 2,657H / 2,864AI & 3,274H / 3,555AI & 2,526H / 2,732AI & 3,146H / 3,464AI \\
Test & 2,403H / 2,814AI & 2,652H / 2,847AI & 3,283H / 3,554AI & 2,535H / 2,742AI & 3,156H / 3,495AI \\
\midrule
\textbf{Split} & \textbf{WP} & \textbf{ROCT} & \textbf{HSwag} & \textbf{SQuAD} & \textbf{SciGen}\\
\midrule
Train & 6,356H / 27,203AI & 3,287H / 27,910AI & 3,129H / 26,882AI & 15,820H / 22,340AI & 4,436H / 21,091AI \\
Val & 3,133H / 3,381AI & 3,284H / 3,483AI & 3,288H / 3,330AI & 2,524H / 2,775AI & 2,531H / 2,612AI \\
Test & 3,099H / 3,437AI & 3,275H / 3,487AI & 3,292H / 3,355AI & 2,508H / 2,796AI & 2,538H / 2,551AI \\
\midrule
\multicolumn{6}{c}{\textbf{Overall Total: Train (93,318H / 249,753AI), Val (28,799H / 30,993AI), Test (28,741H / 31,078AI)}} \\
\bottomrule
\end{tabular}%
}
\caption{MAGE dataset statistics using continuation prompts only. Training includes all available AI generations from 30 models across 7 families (model domains); validation and test sets are approximately balanced (H = Human, AI = AI-generated).}
\label{tab:mage_data_statistics}
\end{table*}

\subsection{Full List of Linguistic Features}
\label{sec:feature-overview}
In the following, we provide an overview of the full set of features used in our experiments. We report features per feature area and rationales for the selection of features wherever we do not use all the available features.

\paragraph{Surface level}
We use the number of tokens, characters, sentences, lemmas, and words over 5 characters, and the raw sequence length in characters (including whitespaces and special characters). We do not consider all additional surface-level features, as we consider them theoretically equivalent to the ones we select.

We do not include the number of types (unique tokens), as it is expected to carry similar information as the number of tokens and the number of lemmas. We do not include sentence-normalized measures, as all information is captured by the combination of the number of sentences and the respective measure to be sentence-normalized. We do not include the average word length, as similar but more interpretable information is captured by the number of long words.

\paragraph{Parts of speech}
We use the number of tokens per universal dependencies POS tag \citep{de-marneffe-etal-2021-universal}, and the POS variability (i.e., the number of different POS tags divided by the number of tokens).

\paragraph{Readability}
We use the total number of syllables in a text, and the Flesch reading ease \citep{kincaid1975derivation} as measures of readability. We do not consider all other readability features, as we consider them theoretically equivalent to the Flesch reading ease or the number of syllables. More specifically, they are all designed to measure the reading level, either measured in school grade levels, age, or a more abstract index of how easy to read a given text is.

\paragraph{Lexical richness}
We use the type-token ratio (TTR), the number of global token hapax legomena (i.e., the number of tokens per text that only occur once across the respective train/evaluation split, computed independently for each split), and the lexical density (i.e., the percentage of adjectives, nouns, verbs, and adverbs per text). We do not consider all other lexical richness features, as we consider them theoretically equivalent to at least one of the selected ones. More specifically, they all measure how many different \textit{kinds} of words relative to the overall number of words are present in a text. Our choice of TTR gives us a general measure, the global token hapax legomena provides information about rare words, and the lexical density about content words.

\paragraph{Information-theoretic}
We use both information-theoretic features, the compressibility and Shannon entropy of a text.

\paragraph{Named entities}
We use the number of named entities per entity type and the number of entities overall in a text.

\paragraph{Semantic}
We use the number of hedges in a text, the average number of synsets overall and per nouns, verbs, adjectives, and adverbs, and the number of tokens with a high and with a low number of synsets overall and per nouns, verbs, adjectives, and adverbs, respectively.

\paragraph{Morphological}
We use all available universal dependencies morphological features that are not uniform in our data.

\paragraph{Syntactic dependencies}
We use the number of universal dependency relations per relation type, and the dependency tree width, tree depth, tree branching factor, the ramification factor, and the number of noun chunks.

\paragraph{Emotion and sentiment}
We use all emotion and sentiment features available in \texttt{elfen}. This includes the number of positive and negative sentiment tokens per text, the number of tokens with a high and a low valence, arousal, and dominance, the average valence, arousal, and dominance, the average intensity, and the number of high and low intensity tokens per Plutchik emotion.

\paragraph{Psycholinguistic}
We use all available psycholinguistic features. This includes the number of tokens with high and low concreteness, age of acquisition, prevalence, socialness, iconicity, and sensorimotor association, and the average rating across tokens per text for these dimensions, respecively.

\subsection{Classifier}
\label{sec:appendix_classifier_details}

We chose a linear Support Vector Machine (SVM) for its effectiveness in high-dimensional feature spaces (284 features), computational efficiency for training hundreds of classifiers across testbeds, and interpretability through feature coefficients.

We employ a linear SVM using scikit-learn's \texttt{SVC} (v1.7.0) implementation\footnote{\url{https://scikit-learn.org/stable/modules/generated/sklearn.svm.SVC.html}} with the following configuration: linear kernel, regularization parameter ($C=1.0$; default value), balanced class weighting to handle class imbalance, random seed 42.
The training was accelerated using Intel Extension for Scikit-learn (\texttt{sklearnex}).

Features were standardized using z-score normalization (zero mean, unit variance) via \texttt{StandardScaler}, fitted on the training set and applied to validation and test sets.

\subsection{Evaluation Metrics}
\label{sec:appendix_metrics_details}

We compute Average Recall (AvgRec) as the arithmetic mean of recall for human-written and AI-generated texts:
\[
\text{AvgRec}
=
\frac{\text{Recall}_{\text{human}} + \text{Recall}_{\text{AI}}}{2}.
\]
Macro F1 is computed by averaging the F1 scores of the human and AI classes. We additionally report accuracy and per-class precision, recall, and F1 where relevant.

\subsection{Testbeds}
\label{sec:testbeds_appendix_details}

We build on the settings designed by \citet{li-etal-2024-mage} (summarized in \Cref{tab:testbeds_config_overview}) and depart from their setup in the following ways.

\paragraph{TB1 (Fixed-domain \& Model-specific)} trained and tested classifiers on individual domain-model pairs, establishing baseline performance for 10 text domains and 30 AI models (300 classifiers).

\subparagraph{TB1-1} extended this setting by adding a family-level analysis: One classifier per text domain-model domain pair (70 classifiers), trained on all generations of a single model family/domain.

\paragraph{TB2 (Arbitrary-text-domains \& Model-domain-specific)} trained on all text domains for each model family/domain separately.

\paragraph{TB3 (Fixed-text-domain \& Arbitrary-model-domains)} evaluated cross-model generalization within individual text domains.

\paragraph{TB4 (Arbitrary-text-domains \& Arbitrary-model-domains)} similar to \citet{li-etal-2024-mage}: combining all available domains and models to assess overall detection capability.

\paragraph{TB5 (Unseen model domains)} evaluated a scenario of leave-one-out by leaving data generated by a specific model family (model domain) out of the training and then evaluated on the excluded model domain.

\paragraph{TB6 (Unseen text domains)} similarly assessed text domain generalization by training on all text domains except one and testing on the excluded one.

\paragraph{TB7 (Unseen text-domains \& Unseen-models)} \citet{li-etal-2024-mage} evaluate on four new text domains and a new model (GPT-4). However, we add GPT-5.6-sol as an unseen model generating data for the same domains and adding a substantially more recent model.

\subparagraph{TB7-1} extended this analysis by evaluating each unseen text domain individually, for each model, to examine text domain-specific performance variations (8 classifiers).

\paragraph{TB8 (Unseen text domain–model domain pairs)} We add a setting excluding both the target text domain and the target model domain from training, giving 70 held-out pairs (10 $\times$ 7). It isolates the joint shift of TB5 and TB6.

\begin{table*}[t]
\centering
\resizebox{\textwidth}{!}{
\begin{tabular}{ll>{\raggedright\arraybackslash}p{3.5cm}>{\raggedright\arraybackslash}p{3.5cm}c}
\toprule
\textbf{TB\#} & \textbf{Name} & \textbf{Training} & \textbf{Testing} & \textbf{\# Classifiers} \\
\midrule
\textbf{TB1} & Fixed-domain \& Model-specific & Single text domain + Single model & Same text domain + Same model & 300 \\
\hline
\textbf{TB1-1} & Fixed-text-domain \& Model-domain-specific & Single text domain + Model domain & Same domain + Same model domain & 70 \\
\hline
\textbf{TB2} & Arbitrary-text-domains \& Model-domain-specific & All text domains + Model domain & All text domains + Same model domain & 7 \\
\hline
\textbf{TB3} & Fixed-text-domain \& Arbitrary-model-domains & Single text domain + All models & Same text domain + All models & 10 \\
\hline
\textbf{TB4} & Arbitrary-text-domains \& Arbitrary-model-domains & All text domains + All models & All text domains + All models & 1 \\
\hline
\textbf{TB5} & Unseen model domains & All text domains + 6 model domains & All domains + 1 held-out model domain & 7 \\
\hline
\textbf{TB6} & Unseen text domains & 9 text domains + All models & 1 held-out text domain + All models & 10 \\
\hline
\textbf{TB7} & Unseen-text \& -model domains & All text domains + All models & 4 new text domains + GPT-4 \& GPT-5.6-sol & 1 \\
\hline
\textbf{TB7-1} & Unseen-text-domain \& Unseen-model & All text domains + All models & 1 of 4 new text domains + 1 of 2 models (GPT4 or GPT-5.6-sol) & 8 \\
\hline
\textbf{TB8} & Unseen text-model-domain pairs & 9 text domains + 6 model domains & 1 held-out text domain + 1 held-out model domain & 70 \\
\bottomrule
\end{tabular}
}
\caption{Overview of experimental testbeds for AI-generated text detection. Each testbed evaluates different generalization scenarios, ranging from fixed conditions (TB1) to completely unseen domain-model combinations (TB7, TB8).}
\label{tab:testbeds_config_overview}
\end{table*}

\subsection{MAGE Baselines Replication}
\label{app:baselines_reimpl}

\citet{li-etal-2024-mage} evaluate three baseline detectors that we use for comparison: FastText, GLTR, and Longformer. Longformer  is trained with their released code and hyperparameters; for FastText and GLTR, no code is released, so we reconstruct the setup from the description in their paper and use library defaults wherever a setting is unspecified. We run each baseline twice: once on the original benchmark to verify our replication against the published numbers, and once on our continuation-only extended data, so that the comparison in \Cref{tab:general_results} reflects differences in the data rather than in the implementation.

\paragraph{FastText} Word bi-grams as the only text features, 100 epochs, all other settings at their defaults. On the original benchmark our replication is within 0.03 AvgRec of the published numbers.

\paragraph{GLTR} GPT-2-XL \citep{radford2019language} gives the number of tokens per text whose rank falls in the Top-10, Top-100, and Top-1000; these are passed to a logistic regression whose solver, penalty and regularization strength are grid-searched under 5-fold cross-validation.

\paragraph{Longformer} \citep{Beltagy2020Longformer} Fine-tuned from \texttt{longformer-base-4096} with MAGE's hyperparameters: sequence length 2048, effective batch size 16, learning rate $3\times10^{-5}$, 5 epochs, mixed precision, seed 42629309. The continuation-only run is distributed over four GPUs with the effective batch size unchanged.

MAGE refines the Longformer decision boundary for the out-of-distribution testbeds; we report the default decision rule unless stated otherwise.

\subsection{Reproducibility Resources}
All experiments were conducted using Python 3.10.18 with the following key dependencies: scikit-learn 1.7.0\footnote{\url{https://scikit-learn.org}} for SVM implementation and evaluation metrics, Intel Extension for Scikit-learn (sklearnex) 2025.9.0\footnote{\url{https://uxlfoundation.github.io/scikit-learn-intelex/}} for accelerated training, pandas 2.3.0\footnote{\url{https://pandas.pydata.org}} for data manipulation, numpy  2.2.6\footnote{\url{https://numpy.org}} for numerical operations, matplotlib 3.10.3\footnote{\url{https://matplotlib.org}} and seaborn 0.13.2\footnote{\url{https://seaborn.pydata.org}} for visualization, spaCy 3.8.7\footnote{\url{https://spacy.io}} with model en\_core\_web\_lg 3.8.0, and ELFEN 1.1.9\footnote{\url{https://github.com/mmmaurer/elfen}} for linguistic feature extraction. More details and required packages can be found in the requirements file (\textit{requirements.txt}) on the repository. All experiments used consistent random seeds (seed=42) for train-test splits and model initialization.

\subsection{Artifact Licensing}
Our code and extensions to the MAGE benchmark are released under the Apache License 2.0, consistent with the license of the original MAGE benchmark. Third-party datasets, model resources, and linguistic resources used in this work remain subject to their respective original licenses and terms of use and are not relicensed by us. We use these artifacts solely for research purposes consistent with their stated access conditions.

%% file: sections/appendix_B.tex
\subsection{Detailed Testbeds Results}

Summary results for all testbeds are reported in \Cref{tab:baseline_aggregated_results_all_testbeds_table}; the main findings are discussed in \Cref{results:detection}.

\begin{table*}[t]
\centering
\resizebox{\textwidth}{!}{%
\begin{tabular}{llccc}
\hline
\textbf{Setting} & \textbf{Method} & \textbf{Acc} & \textbf{AUROC} & \textbf{F1-Macro} \\
\hline
\multicolumn{5}{c}{\textit{In-distribution Detection}} \\
\hline
1. Fixed-Domain \& Model (300 classifiers) & SVM w/ Ling. Feats. & 0.9410 & 0.9893 & 0.7866 \\
1.1. Fixed-Domain \& Model-domain (70 classifiers) & SVM w/ Ling. Feats. & 0.9142 & 0.9884 & 0.8445 \\
2. Arbitrary-Domains \& Fixed-Model (7 classifiers) & SVM w/ Ling. Feats. & 0.8334 & 0.9853 & 0.7407 \\
3. Fixed-Domain \& Arbitrary-Models (10 classifiers) & SVM w/ Ling. Feats. & 0.9055 & 0.9861 & 0.9034 \\
4. Arbitrary-Domains \& Arbitrary-Models (1 classifier) & SVM w/ Ling. Feats. & 0.8348 & 0.9692 & 0.8323 \\
\hline
\multicolumn{5}{c}{\textit{Out-of-distribution Detection}} \\
\hline
5. Unseen Models (7 classifiers) & SVM w/ Ling. Feats. & 0.7294 & 0.9573 & 0.6241 \\
6. Unseen Domains (10 classifiers) & SVM w/ Ling. Feats. & 0.8091 & 0.9640 & 0.8004 \\
7. Unseen Domains \& Unseen Model (1 classifier) & SVM w/ Ling. Feats. & 0.8114 & 0.8829 & 0.8114 \\
7.1. Unseen Domains \& Unseen Models (per domain) (8 classifiers) & SVM w/ Ling. Feats. & 0.8083 & 0.9126 & 0.7892 \\
8. Unseen Domain-Model Pairs (70 classifiers) & SVM w/ Ling. Feats. & 0.6891 & 0.9508 & 0.6024 \\
\hline
\end{tabular}}
\caption{Summary results for all testbeds (TB1-8). For testbeds with multiple classifiers, metrics represent the mean performance averaged across all individual classifiers within that testbed. Results are grouped into \textit{in-distribution} (TB1-4) and \textit{out-of-distribution} (TB5-8) detection scenarios.}
\label{tab:baseline_aggregated_results_all_testbeds_table}
\end{table*}

\begin{table*}[h]
\centering
\resizebox{0.9\textwidth}{!}{%
\begin{tabular}{lccccccccccc}
\toprule
\textbf{Model domain} & \textbf{CMV} & \textbf{ELI5} & \textbf{HSwag} & \textbf{ROCT} & \textbf{SciGen} & \textbf{SQuAD} & \textbf{TLDR} & \textbf{WP} & \textbf{XSum} & \textbf{Yelp} & \textbf{Avg.} \\
\midrule
BigScience & 0.844 & 0.750 & 0.784 & 0.828 & 0.867 & 0.701 & 0.759 & 0.949 & 0.892 & 0.752 & \textbf{0.813} \\
Eleuther & 0.935 & 0.934 & 0.870 & 0.959 & 0.945 & 0.956 & 0.871 & 0.971 & 0.951 & 0.851 & \textbf{0.924} \\
FLAN-T5 & 0.926 & 0.900 & 0.820 & 0.835 & 0.951 & 0.855 & 0.881 & 0.986 & 0.967 & 0.793 & \textbf{0.891} \\
GLM & 0.792 & 0.659 & 0.913 & 0.970 & 0.698 & 0.540 & 0.755 & 0.874 & 0.793 & 0.561 & \textbf{0.756} \\
LLaMA & 0.916 & 0.833 & 0.956 & 0.986 & 0.799 & 0.710 & 0.904 & 0.947 & 0.870 & 0.760 & \textbf{0.868} \\
OpenAI & 0.742 & 0.632 & 0.919 & 0.733 & 0.742 & 0.563 & 0.656 & 0.839 & 0.725 & 0.608 & \textbf{0.716} \\
OPT & 0.960 & 0.911 & 0.951 & 0.964 & 0.967 & 0.936 & 0.945 & 0.978 & 0.966 & 0.858 & \textbf{0.944} \\
\midrule
\textbf{Avg.} & \textbf{0.879} & \textbf{0.801} & \textbf{0.875} & \textbf{0.875} & \textbf{0.852} & \textbf{0.785} & \textbf{0.806} & \textbf{0.939} & \textbf{0.867} & \textbf{0.737} & \textbf{0.842} \\
\bottomrule
\end{tabular}%
}
\caption{TB1.1 results: F1-Macro performance for fixed domain and model domain combinations (70 classifiers total). Each cell represents a separate classifier trained and tested on data from one domain and one model domain. Row averages show model domain detectability; column averages show domain-specific performance.}
\label{tab:baseline_testbed1_modeldomain_fixed_full_features_set}
\end{table*}

\Cref{tab:baseline_testbed1_modeldomain_fixed_full_features_set} reports Macro F1 for TB1.1, reports TB7 results on the four unseen text domains combined and on each domain separately, and \Cref{tab:baseline_testbed8_unseen_pairs_table} reports Macro F1 for all 70 held-out text domain--model domain pairs (TB8).

\begin{table}
\centering
\resizebox{0.85\columnwidth}{!}{%
\begin{tabular}{lcccc}
\toprule
Dataset & \multicolumn{2}{c}{GPT-4} & \multicolumn{2}{c}{GPT-5.6-sol} \\
\cmidrule(lr){2-3} \cmidrule(lr){4-5}
 & AUROC & F1 Macro & AUROC & F1 Macro \\
\midrule
CNN & 0.9986 & 0.9793 & 1.0000 & 0.9925 \\
DialogSum & 0.8275 & 0.4882 & 0.7577 & 0.4431 \\
IMDb & 0.9853 & 0.9048 & 0.8741 & 0.8150 \\
PubMed & 0.9838 & 0.8845 & 0.8735 & 0.8059 \\
\midrule
ALL FOUR & 0.9226 & 0.8278 & 0.8432 & 0.7796 \\
\bottomrule
\end{tabular}}
\caption{Results for TB7: unseen-domains and unseen-models, using the full set of features. The unseen models are GPT-4 and GPT-5.6-sol.}
\label{tab:baseline_testbed7_table}
\end{table}

\begin{table*}[t]
\centering
\resizebox{\textwidth}{!}{%
\begin{tabular}{lccccccccccc}
\hline
\textbf{Model domain} & \textbf{CMV} & \textbf{ELI5} & \textbf{HSwag} & \textbf{ROCT} & \textbf{SciGen} & \textbf{SQuAD} & \textbf{TLDR} & \textbf{WP} & \textbf{XSum} & \textbf{Yelp} & \textbf{Avg.} \\
\hline
BigScience & 0.652 & 0.565 & 0.521 & 0.451 & 0.798 & 0.398 & 0.450 & 0.662 & 0.722 & 0.517 & \textbf{0.574} \\
Eleuther & 0.593 & 0.513 & 0.437 & 0.389 & 0.697 & 0.335 & 0.398 & 0.589 & 0.653 & 0.492 & \textbf{0.510} \\
FLAN-T5 & 0.680 & 0.651 & 0.593 & 0.562 & 0.853 & 0.472 & 0.508 & 0.697 & 0.765 & 0.592 & \textbf{0.637} \\
GLM & 0.642 & 0.534 & 0.466 & 0.403 & 0.779 & 0.389 & 0.436 & 0.639 & 0.669 & 0.505 & \textbf{0.546} \\
LLaMA & 0.783 & 0.685 & 0.529 & 0.510 & 0.847 & 0.500 & 0.563 & 0.839 & 0.840 & 0.638 & \textbf{0.673} \\
OpenAI & 0.699 & 0.594 & 0.507 & 0.466 & 0.834 & 0.438 & 0.483 & 0.627 & 0.706 & 0.579 & \textbf{0.593} \\
OPT & 0.804 & 0.718 & 0.553 & 0.514 & 0.872 & 0.503 & 0.556 & 0.765 & 0.851 & 0.698 & \textbf{0.683} \\
\hline
\textbf{Avg.} & \textbf{0.693} & \textbf{0.608} & \textbf{0.515} & \textbf{0.471} & \textbf{0.811} & \textbf{0.433} & \textbf{0.485} & \textbf{0.688} & \textbf{0.744} & \textbf{0.574} & \textbf{0.602} \\
\hline
\end{tabular}%
}
\caption{Results of TB8; Macro F1 performance on unseen domain-model-domain pairs. Training data excludes both the target domain and target model domain, creating 70 unique domain-model combinations. Each cell shows the F1-Macro score when testing on a domain-model pair that was completely excluded from training. The \textbf{Avg} row shows the average performance across all text domains for each model domain, while the \textbf{Avg} column shows the average performance across all model domains for each domain. Overall statistics: Mean = 0.602, Std = 0.138, Range = [0.335, 0.872].}
\label{tab:baseline_testbed8_unseen_pairs_table}
\end{table*}

\subsection{Cross-Dataset Experiment}
\label{sec:appendix_cmv_cross_dataset_experiment}

To assess prompt generalization, we conducted cross-dataset experiments using the ChangeMyView (CMV) dataset from \citet{donmez-falenska-2025-understand}, which employs direct response prompts rather than continuation prompts. We evaluated: (1) training on MAGE-CMV and testing on their CMV dataset, and (2) training on their dataset and testing on MAGE-CMV. This setting tests whether models can generalize across different prompting strategies for the same domain. Statistics of the final CMV dataset are in \Cref{tab:cmv_data_statistics}.
\Cref{tab:baseline_cmv_full_features_set} reports in-distribution results on CMV.

\begin{table}[t]
    \resizebox{0.8\linewidth}{!}{
    \begin{tabular}{l l c c}
        \toprule
        \textbf{Source} & \textbf{Number of Samples} \\
        \midrule
        Human  & 157,880 comments (13,498 posts) \\
        GPT  & 13,498 (1 comment per post) \\
        LLaMA  & 13,489 (1 comment per post) \\
        Mistral  & 13,498 (1 comment per post)\\
        \bottomrule
    \end{tabular}
    }
    \centering
    \caption{Final dataset sizes for the ChangeMyView (CMV) corpus used in training and evaluation.}
    \label{tab:cmv_data_statistics}
\end{table}

\begin{table}
\centering
\resizebox{0.9\linewidth}{!}{
\begin{tabular}{lccc}
\toprule
AI Text & Accuracy & AUROC & F1 Macro \\
\midrule
\textbf{GPT} & 0.9911 & 0.9994 & 0.9911 \\
\textbf{Llama} & 0.9852 & 0.9948 & 0.9852  \\
\textbf{Mistral} & 0.9920 & 0.9986 & 0.9920  \\
\hline
\textbf{AI-combined} & 0.9884 & 0.9973 & 0.9884 \\
\bottomrule
\end{tabular}}
\caption{Results for \textbf{CMV dataset} for the different settings using the complete features set (284 features). \textit{We also report accuracy since the dataset is balanced.}}
\label{tab:baseline_cmv_full_features_set}
\end{table}

Cross-dataset results are reported in \Cref{tab:cross_cmv_mage_cmv_experiments_1_2,tab:cross_mage_cmv_cmv_experiments_3_4}, respectively. In both scenarios, \textbf{model domain alignment between training and test data has a large impact on detection performance, while prompt formulation appears to have a smaller effect}.

\begin{table*}[t]
\centering
\resizebox{\textwidth}{!}{%
\begin{tabular}{lccccccc}
\hline
Mage Test data & Accuracy & AUROC & F1 Macro & AI-Recall & CMV-Training size & MAGE-CMV-test size \\
\hline
\textbf{i- All models} & 0.7769 & 0.7862 & 0.7687 & 0.5973 & 28,339 H, 28,340 AI & 2,403H, 2,339 AI\\
\textbf{ii-OpenAI-Llama} & 0.9083 & 0.9385 & 0.8725 & 0.7947 & 18,890 H, 18,891 AI & 2,403 H, 750 AI\\
\hline
\end{tabular}}
\caption{Results of two settings for cross-CMV-MAGE setup: (i) classifier is trained on training split from ChangeMyView dataset with all AI-texts (GPT, LLaMA, \& Mistral) and tested on MAGE-CMV data with AI-texts from all 30 models. (ii) classifier trained on training split from ChangeMyView but with AI-texts from only GPT \& LLaMA, then tested on MAGE-CMV with AI-texts from OpenAI \& LLaMA models.}
\label{tab:cross_cmv_mage_cmv_experiments_1_2}
\end{table*}

\begin{table*}[t]
\centering
\resizebox{\textwidth}{!}{%
\begin{tabular}{lccccccc}
\hline
Mage Test data & Accuracy & AUROC & F1 Macro & AI-Recall & MAGE-CMV-Training size & CMV-test size \\
\hline
\textbf{i- All models} & 0.4796 & 0.5529 & 0.4528 & 0.2582 & 4,223 H, 18,956 AI & 8,097 H, 8,097 AI\\
\textbf{ii-OpenAI-Llama} & 0.8496 & 0.9121 & 0.8492 & 0.9016 & 4,223 H, 6,029 AI & 5,398 H, 5,397 AI\\
\hline
\end{tabular}
}
\caption{Results of two settings for cross-CMV-MAGE setup: (i) classifier is trained on training split from MAGE-CMV data with AI-texts from all 30 models and tested on ChangeMyView dataset with all AI-texts (GPT, LLaMA, \& Mistral). (ii) classifier trained on training split from MAGE-CMV with AI-texts from OpenAI \& LLaMA models, then tested on ChangeMyView but with AI-texts from only GPT \& LLaMA.}
\label{tab:cross_mage_cmv_cmv_experiments_3_4}
\end{table*}

%% file: sections/appendix_C.tex
In this section, we discuss further the ablation studies conducted on all testbeds (ablation study \textbf{A}) and on domain-agnostic settings (ablation study \textbf{B} -cumulative).

\subsection{Model and Text Domain Effects Additional Analysis}
\label{sec:appendix_ablation_testbeds_2356}
\begin{figure*}[t]

    \centering
    \includegraphics[width=0.95\linewidth]{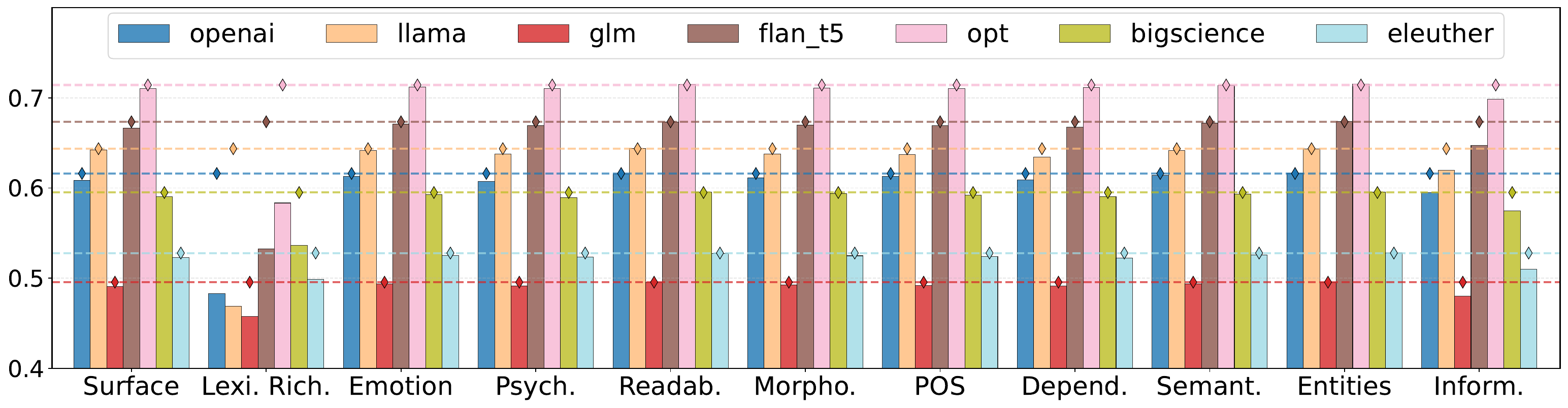}
    \caption{Visualization of results of ablation study on \textbf{TB5} (unseen models). The horizontal dashed lines indicate the original baseline results for each model, and the bars indicate the results of dropping the corresponding feature area.}
    \label{fig:ablation_results_testbed5_figure_app}
\end{figure*}

\begin{figure*}[t]
    \centering
    \begin{subfigure}[t]{0.48\textwidth}
        \centering
        \includegraphics[width=\linewidth]{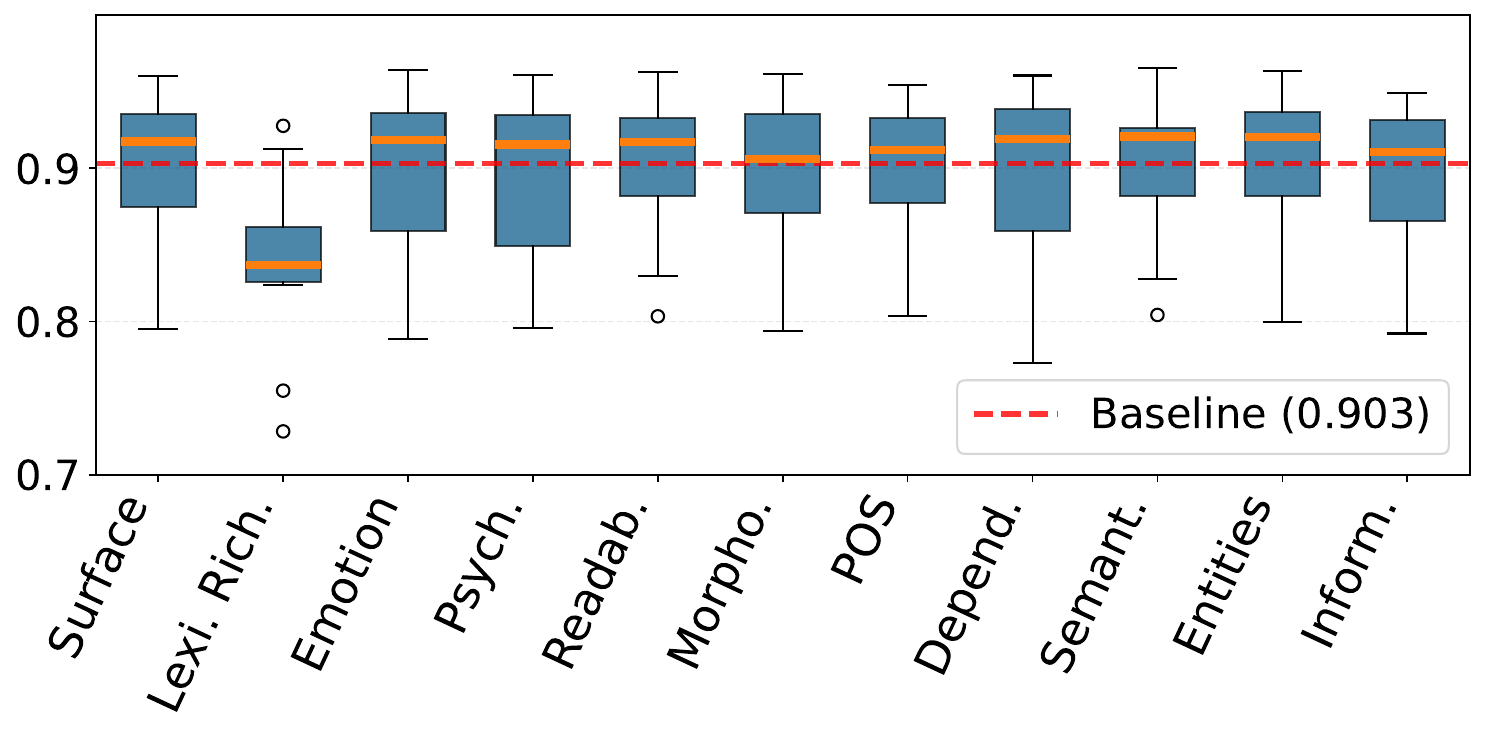}
        \caption{Visualization of results distribution of ablation study on \textbf{TB3} (domain-fixed—arbitrary-model-domains) across the 11 feature areas.}
        \label{fig:ablation_results_testbed3_box_plot}
    \end{subfigure}
    \hfill
    \begin{subfigure}[t]{0.48\textwidth}
        \centering
        \includegraphics[width=\linewidth]{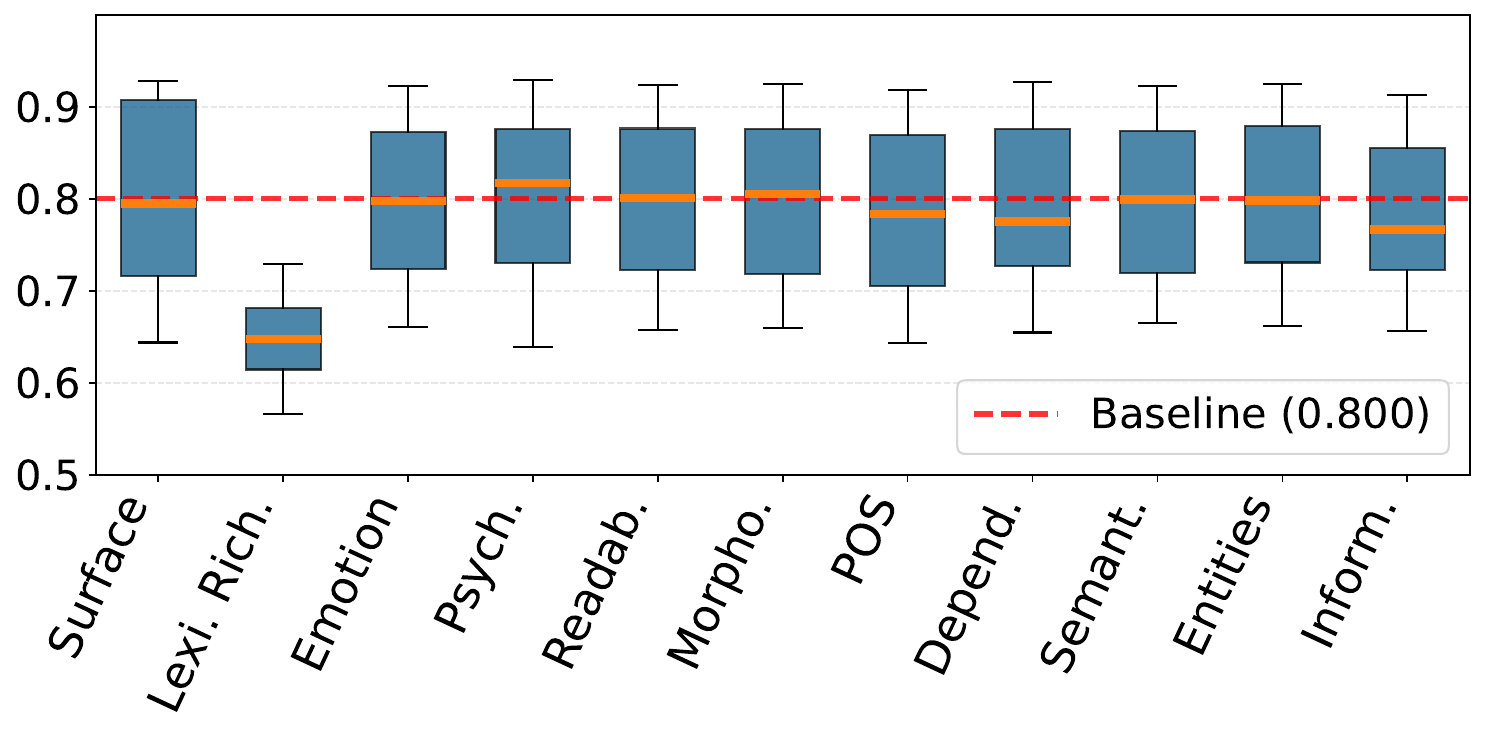}
        \caption{Visualization of results distribution of ablation study on \textbf{TB6} (unseen domains) across the 11 feature areas.}
        \label{fig:ablation_results_testbed6_box_plot}
    \end{subfigure}
    \caption{Distribution of F1-Macro scores across feature area ablations for in-distribution (TB3) and out-of-distribution (TB6) testbeds focusing on text domains.}
    \label{fig:ablation_testbed3_6_comparison}
\end{figure*}

\Cref{tab:ablation} reports the complete ablation results for TB2, TB3, TB5, and TB6. \Cref{fig:ablation_results_testbed5_figure_app} visualizes the TB5 results, and \Cref{fig:ablation_testbed3_6_comparison} shows the distribution of TB3 and TB6 results across feature-area ablations.

\paragraph{Model domain Effects (TB2 \& TB5)}
The upper part of \Cref{tab:ablation} reports results for TB2 (ID) and TB5 (OOD).

\paragraph{Text Domain Effects (TB3 \& TB6)}
The lower part of \Cref{tab:ablation} reports results for TB3 (ID) and TB6 (OOD), and \Cref{fig:ablation_testbed3_6_comparison} shows their distributions.

\subsection{Zoom-in TB7: Unseen Domains and Unseen Model}
\label{sec:testebed7_appendix_unseendomains_unseenmodels}

For the setting of completely unseen text and model domains, we tested two configurations: combining all four unseen domains (CNN, PubMed, IMDb, DialogSum) into a single test set, and testing each domain separately and per model.

\paragraph{Combined Unseen Domains}
Results for the four unseen text domains combined are reported in \Cref{tab:ablation_testbed4_results_horizontal} and discussed in \Cref{results:ablation}.

\begin{figure}[t]
    \centering
    \includegraphics[width=1\linewidth]{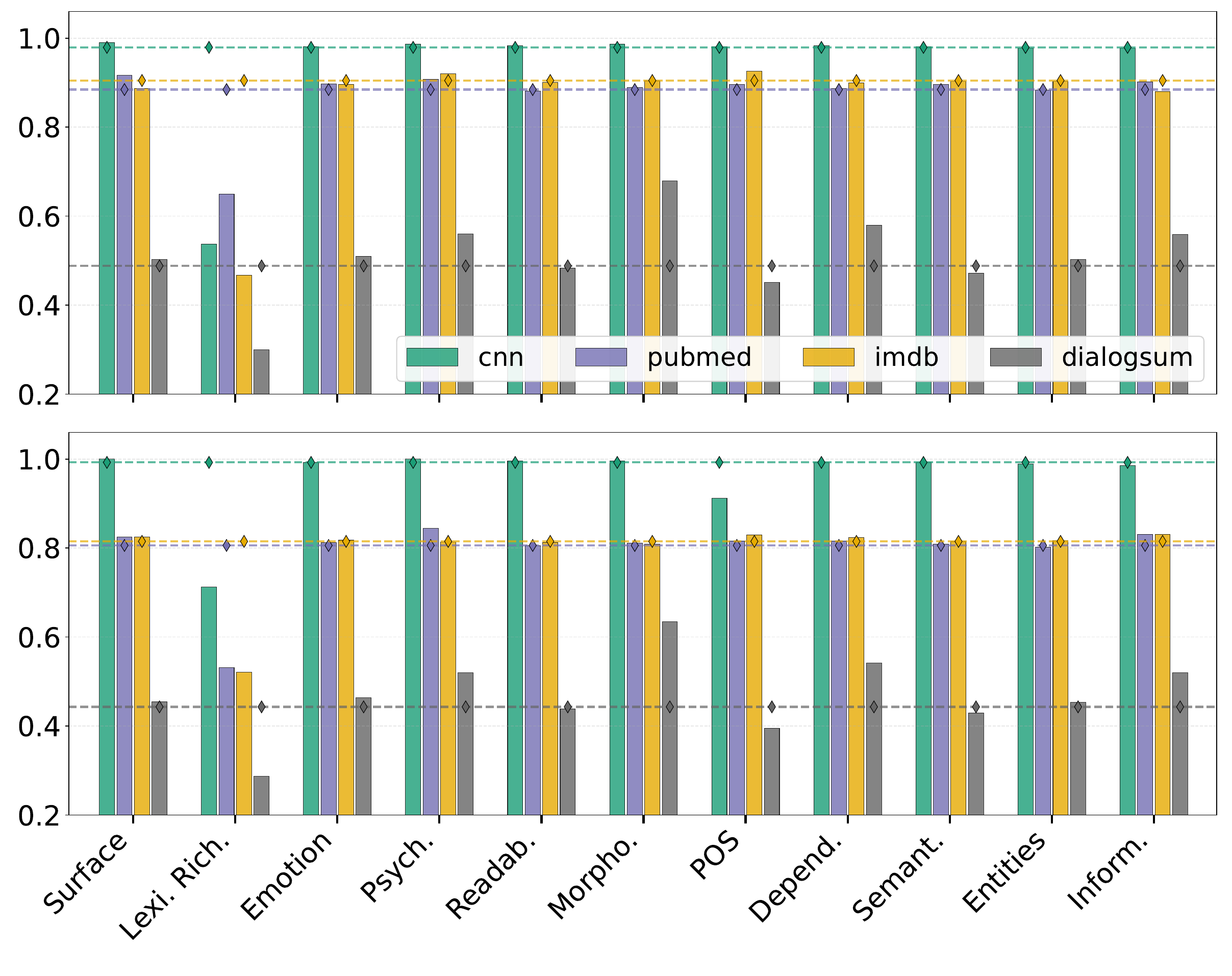}
    \caption{Visualization of results of ablation study on \textbf{TB7} using separate unseen domains (cnn, pubmed, imdb, \& dialogsum) and two unseen models, shown in separate panels: GPT-4 (top) and GPT-5.6-sol (bottom). The panels are separated so that the effect of each ablated feature area can be compared across a model seen in the original MAGE data and a substantially more recent one. The horizontal dashed lines indicate the original baseline results for each domain, and the bars indicate the results of dropping the corresponding feature area.}
    \label{fig:ablation_results_testbed7_separate_domains_figure}
\end{figure}

\begin{table*}[t]
\centering
\resizebox{\textwidth}{!}{
\begin{tabular}{lcccccccccccc}
\toprule
Domain & Baseline & Surface & Lexi.Rich. & Emotion & Psych. & Readab. & Morpho. & POS & Depend. & Semant. & Entities & Inform. \\
\midrule
\multicolumn{13}{c}{\textbf{Testbed 2}} \\
\midrule
OpenAI & .555 & .551 {\color{red}(-.005)} & .558 {\color{green!60!black}(+.003)} & .554 {\color{red}(-.002)} & .549 {\color{red}(-.007)} & .554 {\color{red}(-.002)} & .554 {\color{red}(-.001)} & .557 {\color{green!60!black}(+.001)} & .549 {\color{red}(-.006)} & .556 {\color{green!60!black}(+.000)} & .556 {\color{green!60!black}(+.001)} & .552 {\color{red}(-.003)} \\
LLaMA & .786 & .782 {\color{red}(-.005)} & .779 {\color{red}(-.008)} & .787 {\color{green!60!black}(+.001)} & .787 {\color{green!60!black}(+.001)} & .787 {\color{green!60!black}(+.001)} & .781 {\color{red}(-.005)} & .769 {\color{red}(-.017)} & .784 {\color{red}(-.003)} & .786 {\color{green!60!black}(+.000)} & .786 {\color{green!60!black}(+.000)} & .785 {\color{red}(-.001)} \\
GLM & .627 & .613 {\color{red}(-.014)} & .643 {\color{green!60!black}(+.016)} & .626 {\color{red}(-.001)} & .618 {\color{red}(-.009)} & .628 {\color{green!60!black}(+.001)} & .621 {\color{red}(-.006)} & .610 {\color{red}(-.017)} & .622 {\color{red}(-.005)} & .628 {\color{green!60!black}(+.001)} & .627 {\color{red}(-.000)} & .626 {\color{red}(-.001)} \\
FLAN-T5 & .766 & .752 {\color{red}(-.014)} & .671 {\color{red}(-.095)} & .765 {\color{red}(-.001)} & .761 {\color{red}(-.004)} & .765 {\color{red}(-.001)} & .757 {\color{red}(-.009)} & .765 {\color{red}(-.001)} & .762 {\color{red}(-.004)} & .764 {\color{red}(-.002)} & .765 {\color{red}(-.001)} & .763 {\color{red}(-.002)} \\
OPT & .913 & .906 {\color{red}(-.007)} & .881 {\color{red}(-.032)} & .913 {\color{green!60!black}(+.000)} & .910 {\color{red}(-.003)} & .913 {\color{green!60!black}(+.000)} & .910 {\color{red}(-.003)} & .911 {\color{red}(-.002)} & .909 {\color{red}(-.004)} & .913 {\color{red}(-.000)} & .913 {\color{green!60!black}(+.000)} & .909 {\color{red}(-.004)} \\
BigScience & .656 & .646 {\color{red}(-.010)} & .599 {\color{red}(-.057)} & .654 {\color{red}(-.002)} & .647 {\color{red}(-.009)} & .657 {\color{green!60!black}(+.001)} & .634 {\color{red}(-.022)} & .651 {\color{red}(-.005)} & .650 {\color{red}(-.006)} & .654 {\color{red}(-.002)} & .655 {\color{red}(-.002)} & .649 {\color{red}(-.007)} \\
Eleuther & .881 & .875 {\color{red}(-.006)} & .857 {\color{red}(-.024)} & .882 {\color{green!60!black}(+.001)} & .878 {\color{red}(-.003)} & .881 {\color{green!60!black}(+.000)} & .870 {\color{red}(-.011)} & .877 {\color{red}(-.004)} & .876 {\color{red}(-.005)} & .879 {\color{red}(-.002)} & .881 {\color{red}(-.000)} & .878 {\color{red}(-.003)} \\
\midrule
\multicolumn{13}{c}{\textbf{Testbed 5}} \\
\midrule
OpenAI & .610 & .605 {\color{red}(-.005)} & .489 {\color{red}(-.121)} & .610 {\color{red}(-.000)} & .607 {\color{red}(-.004)} & .610 {\color{green!60!black}(+.000)} & .607 {\color{red}(-.003)} & .593 {\color{red}(-.017)} & .607 {\color{red}(-.003)} & .610 {\color{green!60!black}(+.000)} & .611 {\color{green!60!black}(+.001)} & .593 {\color{red}(-.017)} \\
LLaMA & .692 & .691 {\color{red}(-.001)} & .512 {\color{red}(-.180)} & .690 {\color{red}(-.002)} & .689 {\color{red}(-.004)} & .693 {\color{green!60!black}(+.001)} & .688 {\color{red}(-.005)} & .682 {\color{red}(-.010)} & .687 {\color{red}(-.005)} & .691 {\color{red}(-.001)} & .692 {\color{green!60!black}(+.000)} & .673 {\color{red}(-.019)} \\
GLM & .562 & .558 {\color{red}(-.004)} & .488 {\color{red}(-.073)} & .561 {\color{red}(-.001)} & .558 {\color{red}(-.004)} & .562 {\color{green!60!black}(+.000)} & .559 {\color{red}(-.003)} & .548 {\color{red}(-.014)} & .558 {\color{red}(-.003)} & .560 {\color{red}(-.002)} & .562 {\color{green!60!black}(+.000)} & .544 {\color{red}(-.017)} \\
FLAN-T5 & .670 & .666 {\color{red}(-.004)} & .525 {\color{red}(-.144)} & .670 {\color{red}(-.000)} & .668 {\color{red}(-.001)} & .670 {\color{green!60!black}(+.000)} & .666 {\color{red}(-.004)} & .651 {\color{red}(-.019)} & .666 {\color{red}(-.004)} & .669 {\color{red}(-.001)} & .670 {\color{green!60!black}(+.000)} & .645 {\color{red}(-.024)} \\
OPT & .715 & .712 {\color{red}(-.003)} & .579 {\color{red}(-.136)} & .713 {\color{red}(-.001)} & .712 {\color{red}(-.003)} & .715 {\color{green!60!black}(+.000)} & .713 {\color{red}(-.002)} & .696 {\color{red}(-.019)} & .712 {\color{red}(-.002)} & .715 {\color{green!60!black}(+.000)} & .715 {\color{green!60!black}(+.000)} & .702 {\color{red}(-.012)} \\
BigScience & .594 & .590 {\color{red}(-.004)} & .536 {\color{red}(-.057)} & .592 {\color{red}(-.001)} & .591 {\color{red}(-.003)} & .594 {\color{green!60!black}(+.000)} & .592 {\color{red}(-.002)} & .577 {\color{red}(-.016)} & .591 {\color{red}(-.003)} & .593 {\color{red}(-.001)} & .593 {\color{red}(-.000)} & .575 {\color{red}(-.019)} \\
Eleuther & .527 & .523 {\color{red}(-.003)} & .497 {\color{red}(-.029)} & .524 {\color{red}(-.002)} & .524 {\color{red}(-.003)} & .527 {\color{green!60!black}(+.000)} & .524 {\color{red}(-.002)} & .512 {\color{red}(-.014)} & .523 {\color{red}(-.004)} & .525 {\color{red}(-.001)} & .526 {\color{red}(-.000)} & .511 {\color{red}(-.015)} \\
\midrule
\multicolumn{13}{c}{\textbf{Testbed 3}} \\
\midrule
CMV & .925 & .921 {\color{red}(-.004)} & .829 {\color{red}(-.096)} & .923 {\color{red}(-.002)} & .914 {\color{red}(-.011)} & .926 {\color{green!60!black}(+.002)} & .923 {\color{red}(-.002)} & .908 {\color{red}(-.016)} & .919 {\color{red}(-.006)} & .926 {\color{green!60!black}(+.001)} & .925 {\color{green!60!black}(+.000)} & .911 {\color{red}(-.014)} \\
Yelp & .803 & .800 {\color{red}(-.003)} & .717 {\color{red}(-.086)} & .790 {\color{red}(-.013)} & .799 {\color{red}(-.004)} & .803 {\color{red}(-.000)} & .791 {\color{red}(-.013)} & .797 {\color{red}(-.006)} & .788 {\color{red}(-.015)} & .801 {\color{red}(-.002)} & .799 {\color{red}(-.004)} & .791 {\color{red}(-.012)} \\
XSum & .945 & .943 {\color{red}(-.002)} & .834 {\color{red}(-.111)} & .942 {\color{red}(-.003)} & .943 {\color{red}(-.002)} & .945 {\color{green!60!black}(+.000)} & .943 {\color{red}(-.002)} & .939 {\color{red}(-.007)} & .948 {\color{green!60!black}(+.002)} & .946 {\color{green!60!black}(+.001)} & .945 {\color{green!60!black}(+.000)} & .938 {\color{red}(-.007)} \\
TLDR & .916 & .913 {\color{red}(-.002)} & .817 {\color{red}(-.099)} & .915 {\color{red}(-.001)} & .921 {\color{green!60!black}(+.006)} & .915 {\color{red}(-.000)} & .905 {\color{red}(-.011)} & .915 {\color{red}(-.001)} & .910 {\color{red}(-.006)} & .915 {\color{red}(-.001)} & .918 {\color{green!60!black}(+.002)} & .915 {\color{red}(-.001)} \\
ELI5 & .876 & .870 {\color{red}(-.006)} & .748 {\color{red}(-.128)} & .873 {\color{red}(-.003)} & .871 {\color{red}(-.005)} & .877 {\color{green!60!black}(+.001)} & .868 {\color{red}(-.008)} & .866 {\color{red}(-.010)} & .864 {\color{red}(-.012)} & .877 {\color{green!60!black}(+.001)} & .876 {\color{green!60!black}(+.000)} & .856 {\color{red}(-.020)} \\
WP & .964 & .962 {\color{red}(-.002)} & .901 {\color{red}(-.063)} & .965 {\color{green!60!black}(+.000)} & .962 {\color{red}(-.003)} & .964 {\color{green!60!black}(+.000)} & .962 {\color{red}(-.002)} & .955 {\color{red}(-.009)} & .962 {\color{red}(-.002)} & .964 {\color{red}(-.000)} & .964 {\color{green!60!black}(+.000)} & .950 {\color{red}(-.015)} \\
ROCT & .931 & .932 {\color{green!60!black}(+.001)} & .873 {\color{red}(-.059)} & .935 {\color{green!60!black}(+.004)} & .922 {\color{red}(-.009)} & .934 {\color{green!60!black}(+.002)} & .921 {\color{red}(-.011)} & .934 {\color{green!60!black}(+.003)} & .930 {\color{red}(-.001)} & .933 {\color{green!60!black}(+.001)} & .930 {\color{red}(-.001)} & .936 {\color{green!60!black}(+.004)} \\
HSwag & .957 & .955 {\color{red}(-.002)} & .931 {\color{red}(-.026)} & .959 {\color{green!60!black}(+.002)} & .950 {\color{red}(-.007)} & .956 {\color{red}(-.001)} & .952 {\color{red}(-.004)} & .952 {\color{red}(-.004)} & .954 {\color{red}(-.003)} & .956 {\color{red}(-.000)} & .955 {\color{red}(-.002)} & .950 {\color{red}(-.007)} \\
SQuAD & .803 & .801 {\color{red}(-.002)} & .801 {\color{red}(-.002)} & .798 {\color{red}(-.005)} & .800 {\color{red}(-.003)} & .801 {\color{red}(-.002)} & .800 {\color{red}(-.003)} & .794 {\color{red}(-.009)} & .802 {\color{red}(-.001)} & .800 {\color{red}(-.003)} & .804 {\color{green!60!black}(+.001)} & .802 {\color{red}(-.001)} \\
SciGen & .913 & .912 {\color{red}(-.001)} & .818 {\color{red}(-.095)} & .914 {\color{green!60!black}(+.001)} & .914 {\color{green!60!black}(+.001)} & .913 {\color{red}(-.000)} & .916 {\color{green!60!black}(+.003)} & .906 {\color{red}(-.007)} & .912 {\color{red}(-.001)} & .914 {\color{green!60!black}(+.001)} & .916 {\color{green!60!black}(+.003)} & .905 {\color{red}(-.008)} \\
\midrule
\multicolumn{13}{c}{\textbf{Testbed 6}} \\
\midrule
CMV & .875 & .882 {\color{green!60!black}(+.006)} & .729 {\color{red}(-.146)} & .873 {\color{red}(-.002)} & .886 {\color{green!60!black}(+.011)} & .878 {\color{green!60!black}(+.002)} & .857 {\color{red}(-.019)} & .858 {\color{red}(-.018)} & .879 {\color{green!60!black}(+.004)} & .868 {\color{red}(-.008)} & .876 {\color{green!60!black}(+.000)} & .836 {\color{red}(-.040)} \\
Yelp & .783 & .787 {\color{green!60!black}(+.004)} & .657 {\color{red}(-.126)} & .775 {\color{red}(-.007)} & .801 {\color{green!60!black}(+.018)} & .785 {\color{green!60!black}(+.003)} & .786 {\color{green!60!black}(+.003)} & .770 {\color{red}(-.013)} & .776 {\color{red}(-.007)} & .780 {\color{red}(-.003)} & .780 {\color{red}(-.003)} & .755 {\color{red}(-.027)} \\
XSum & .910 & .929 {\color{green!60!black}(+.019)} & .631 {\color{red}(-.278)} & .908 {\color{red}(-.002)} & .924 {\color{green!60!black}(+.014)} & .913 {\color{green!60!black}(+.003)} & .925 {\color{green!60!black}(+.015)} & .888 {\color{red}(-.022)} & .927 {\color{green!60!black}(+.017)} & .911 {\color{green!60!black}(+.001)} & .907 {\color{red}(-.003)} & .870 {\color{red}(-.039)} \\
TLDR & .713 & .687 {\color{red}(-.026)} & .609 {\color{red}(-.104)} & .720 {\color{green!60!black}(+.007)} & .703 {\color{red}(-.010)} & .716 {\color{green!60!black}(+.003)} & .712 {\color{red}(-.001)} & .708 {\color{red}(-.005)} & .704 {\color{red}(-.009)} & .711 {\color{red}(-.002)} & .725 {\color{green!60!black}(+.012)} & .721 {\color{green!60!black}(+.008)} \\
ELI5 & .819 & .803 {\color{red}(-.015)} & .689 {\color{red}(-.130)} & .820 {\color{green!60!black}(+.002)} & .833 {\color{green!60!black}(+.015)} & .817 {\color{red}(-.002)} & .824 {\color{green!60!black}(+.006)} & .797 {\color{red}(-.022)} & .774 {\color{red}(-.044)} & .818 {\color{red}(-.000)} & .817 {\color{red}(-.002)} & .778 {\color{red}(-.041)} \\
WP & .878 & .916 {\color{green!60!black}(+.038)} & .660 {\color{red}(-.218)} & .871 {\color{red}(-.007)} & .846 {\color{red}(-.032)} & .873 {\color{red}(-.005)} & .883 {\color{green!60!black}(+.005)} & .874 {\color{red}(-.004)} & .866 {\color{red}(-.012)} & .875 {\color{red}(-.003)} & .880 {\color{green!60!black}(+.002)} & .862 {\color{red}(-.015)} \\
ROCT & .698 & .706 {\color{green!60!black}(+.008)} & .567 {\color{red}(-.132)} & .711 {\color{green!60!black}(+.012)} & .726 {\color{green!60!black}(+.028)} & .693 {\color{red}(-.006)} & .710 {\color{green!60!black}(+.012)} & .705 {\color{green!60!black}(+.007)} & .711 {\color{green!60!black}(+.013)} & .699 {\color{green!60!black}(+.000)} & .679 {\color{red}(-.020)} & .728 {\color{green!60!black}(+.030)} \\
HSwag & .745 & .746 {\color{green!60!black}(+.001)} & .568 {\color{red}(-.177)} & .734 {\color{red}(-.011)} & .744 {\color{red}(-.001)} & .745 {\color{green!60!black}(+.000)} & .737 {\color{red}(-.008)} & .656 {\color{red}(-.088)} & .775 {\color{green!60!black}(+.030)} & .745 {\color{green!60!black}(+.000)} & .749 {\color{green!60!black}(+.004)} & .696 {\color{red}(-.049)} \\
SQuAD & .658 & .644 {\color{red}(-.014)} & .639 {\color{red}(-.020)} & .661 {\color{green!60!black}(+.003)} & .639 {\color{red}(-.019)} & .658 {\color{red}(-.001)} & .659 {\color{green!60!black}(+.001)} & .643 {\color{red}(-.015)} & .655 {\color{red}(-.004)} & .665 {\color{green!60!black}(+.006)} & .662 {\color{green!60!black}(+.004)} & .656 {\color{red}(-.002)} \\
SciGen & .925 & .919 {\color{red}(-.006)} & .704 {\color{red}(-.221)} & .922 {\color{red}(-.003)} & .929 {\color{green!60!black}(+.004)} & .924 {\color{red}(-.002)} & .923 {\color{red}(-.002)} & .918 {\color{red}(-.007)} & .923 {\color{red}(-.002)} & .923 {\color{red}(-.002)} & .925 {\color{red}(-.000)} & .913 {\color{red}(-.012)} \\
\bottomrule
\end{tabular}}
\caption{Complete results of the ablation study (A) where we drop a feature group at each training cycle for four different testbeds (\textbf{TB2}, \textbf{TB3}, \textbf{TB5}, \& \textbf{TB6}) and the difference to baseline where the classifier is trained on the full set of features (284). The values in red indicate a decrease in performance, while green values indicate an improvement compared to the baseline result.}
\label{tab:ablation}
\end{table*}

\paragraph{Separate Unseen Domains}
\Cref{fig:ablation_results_testbed7_separate_domains_figure} reports the ablation results for each unseen text domain separately, for GPT-4 (top) and GPT-5.6-sol (bottom).

\subsection{TB8 Ablation: Selection of Domain-Model Pairs}
\label{sec:testebed8_selection_16_pairs}
Running ablation on all TB8 combinations (10 text domains × 7 model domains × 11 feature areas = 770 classifiers) is computationally expensive. We selected 16 representative pairs based on sensitivity analysis.

\begin{figure*}[t]
    \centering
    \includegraphics[width=0.8\linewidth]{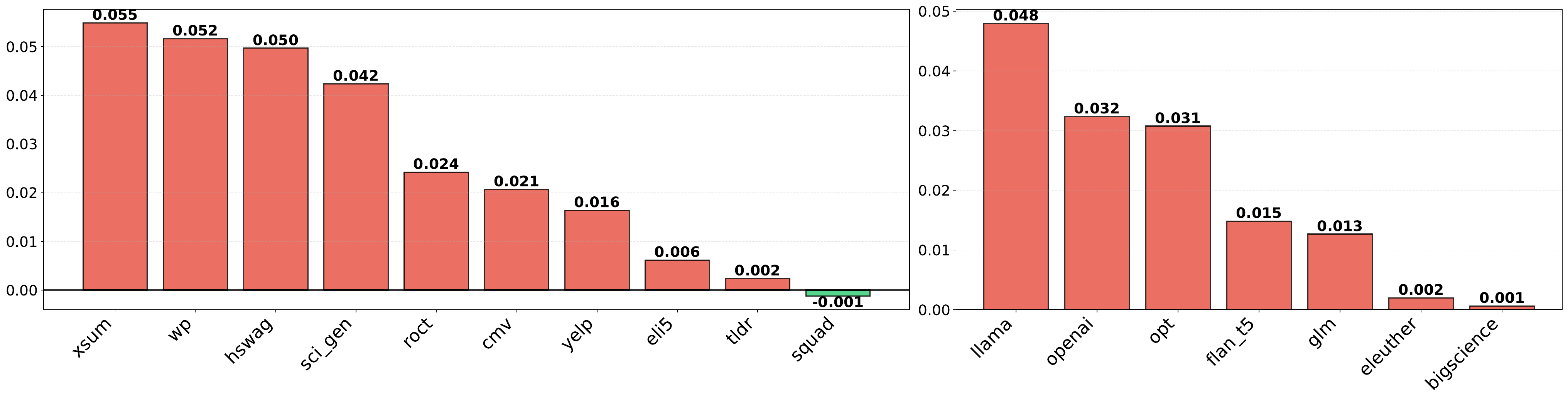}
    \caption{Change in standard deviation between unseen (\textbf{TB5} \& \textbf{TB6}) and fixed (\textbf{TB2} \& \textbf{TB3}) scenarios for text domains (left) and model domains (right). Red bars indicate increased sensitivity when unseen, green bars indicate decreased sensitivity. Higher absolute values represent larger shifts in model behavior across different experimental setups.}
    \label{fig:difference_std_unseen_vs_fixed}
\end{figure*}

\begin{figure}[t]
    \centering
    \includegraphics[width=1\linewidth]{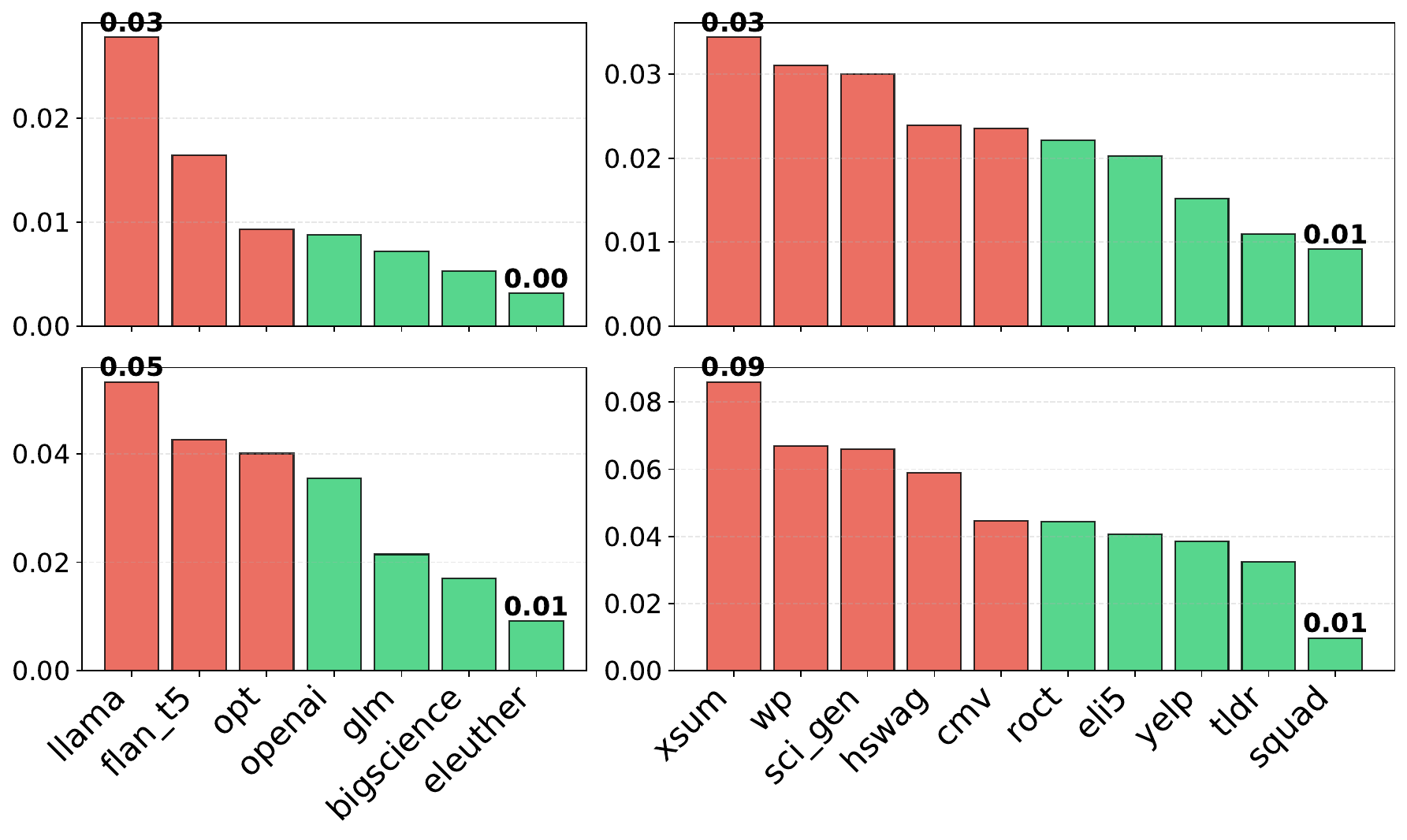}
    \caption{Results of the standard deviation between ablation study across the different testbeds and the ablation study using feature areas. Red bars indicate increased sensitivity compared to the mean, green bars indicate decreased sensitivity compared to the mean (horizontal line).}
    \label{fig:standard_deviation_results}
\end{figure}

\paragraph{Model and Domain Selection}
We measure how much the feature-area ablation effects vary (standard deviation across the 11 ablations) for each text and model domain in the ID testbeds (TB2, TB3) and the OOD testbeds (TB5, TB6) (\Cref{fig:standard_deviation_results}), and compute the change from ID to OOD (\Cref{fig:difference_std_unseen_vs_fixed}).

\paragraph{Selected 4 text domains:} XSum and WritingPrompts (the largest increases in standard deviation from ID to OOD), and TLDR and SQuAD (almost no change).

\paragraph{Selected 4 model domains:} LLaMA and OpenAI (the largest increases in standard deviation from ID to OOD), and EleutherAI and BigScience (almost no change).

This yields 16 pairs covering the full spectrum from stable (SQuAD-EleutherAI) to volatile (XSum-LLaMA).

\begin{figure}
    \centering
    \includegraphics[width=1\linewidth]{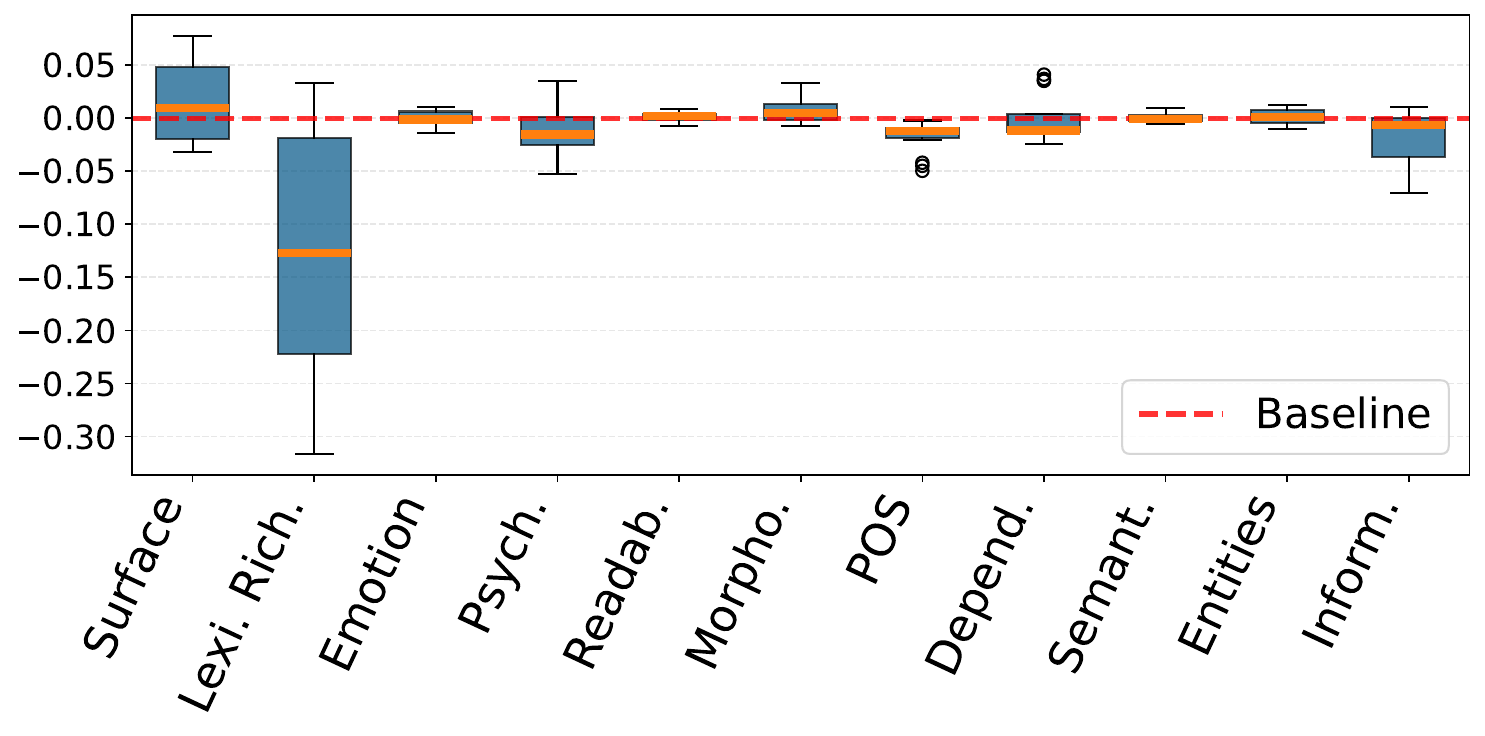}
    \caption{Leave-one-out features-area ablation performance variance across 16 text domain-model domain}
    \label{fig:ablation_study_for_16_domain_model_pair}
\end{figure}

\begin{figure}[t]
    \centering
    \includegraphics[width=1\linewidth]{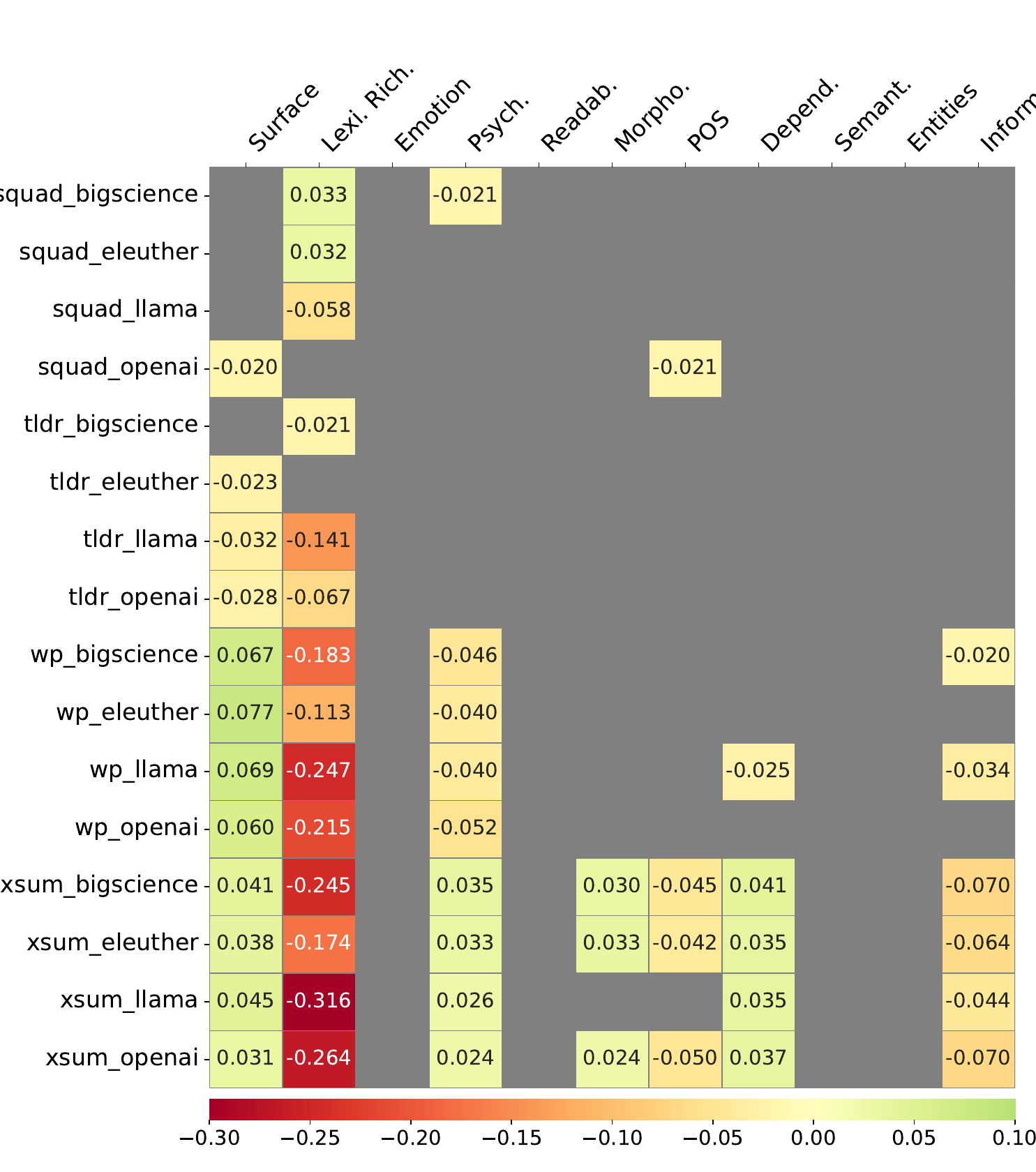}
    \caption{The difference in the performance for the 16 selected text domain-model-domain pairs from the previous study. We keep only the deltas ($\Delta = ablation\_results - baseline$) of $\pm 2\%$ effect.}
    \label{fig:detailed_results_of_ablation_for_16_domain_domain_pair}
\end{figure}

\begin{figure}[t]
    \centering
    \includegraphics[width=1\linewidth]{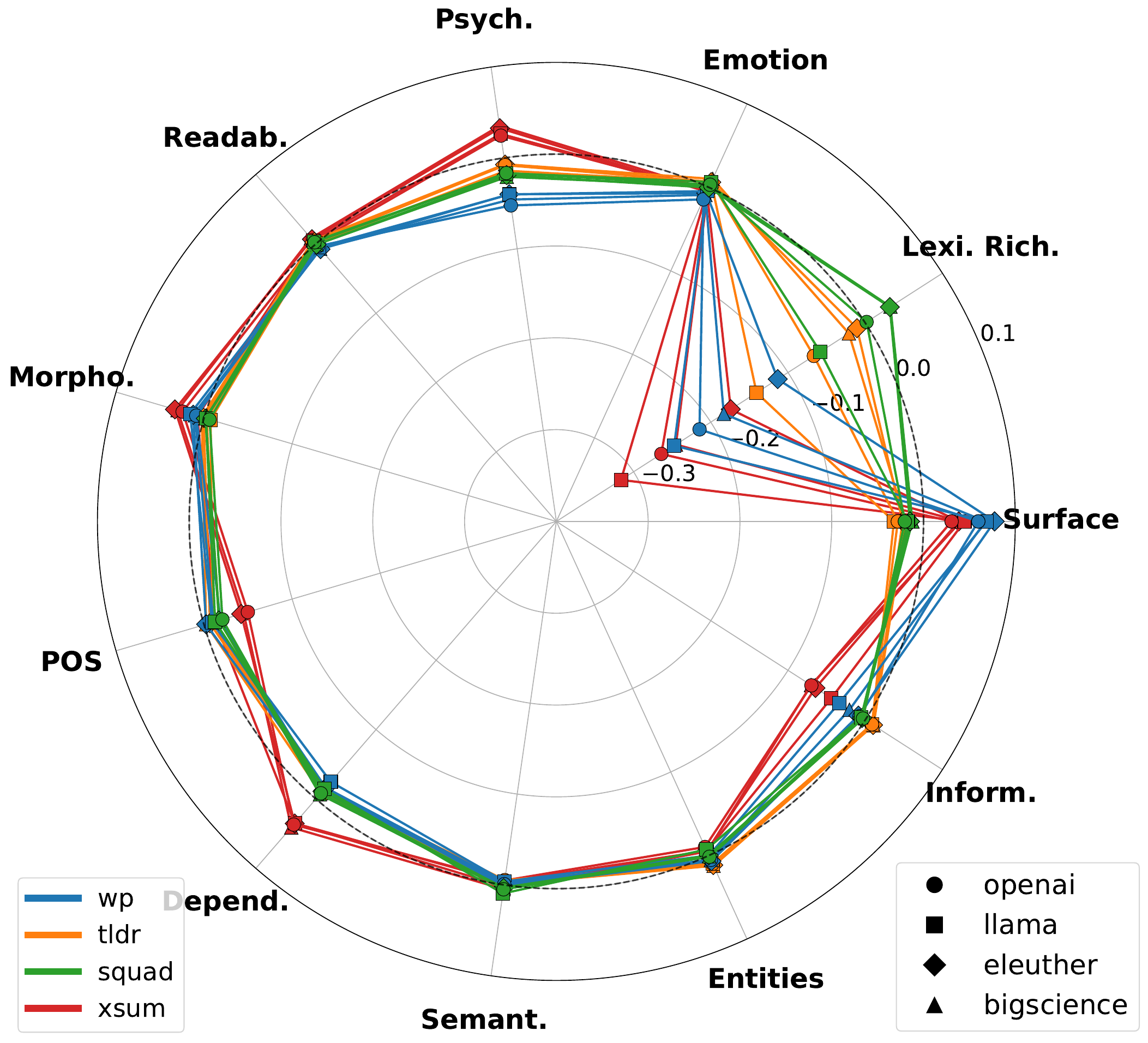}
    \caption{The difference in the performance for the 16 selected text-domain-model-domain pairs from the previous study using the 16 text-domain-model-domain pairs. The text domains separated by colors and and the model domains by markers/ shapes.}
    \label{fig:spiderweb_plot_ablation_for_16_domain_domain_pairs_app}
\end{figure}

\paragraph{Results}
\Cref{fig:ablation_study_for_16_domain_model_pair} shows the distribution of ablation effects across the 16 pairs, \Cref{fig:detailed_results_of_ablation_for_16_domain_domain_pair} reports all effects of at least 2 percentage points per pair, and \Cref{fig:spiderweb_plot_ablation_for_16_domain_domain_pairs_app} summarizes the patterns per feature area.

Overall, we observe that the text domain largely determines which feature areas matter, while the model domain modulates the size, and in some cases the direction, of these effects.

\subsection{Lexical Richness Results in Unseen Domain-Model Pairs (TB8)}
\label{sec:lexical_richness_results_testbed8}
Our ablation studies identified lexical richness as the most influential feature area. In the cumulative ablation on TB7, a classifier using \textit{only} lexical richness features outperformed the full-feature baseline (\Cref{fig:ablation_cumulative_testbed7}). This motivated us to test whether the same holds for individual unseen domain-model pairs in TB8.

\begin{figure*}[t]
  \centering
  \begin{subfigure}[t]{0.48\textwidth}\vspace{0pt}
    \centering
    \includegraphics[width=\textwidth]{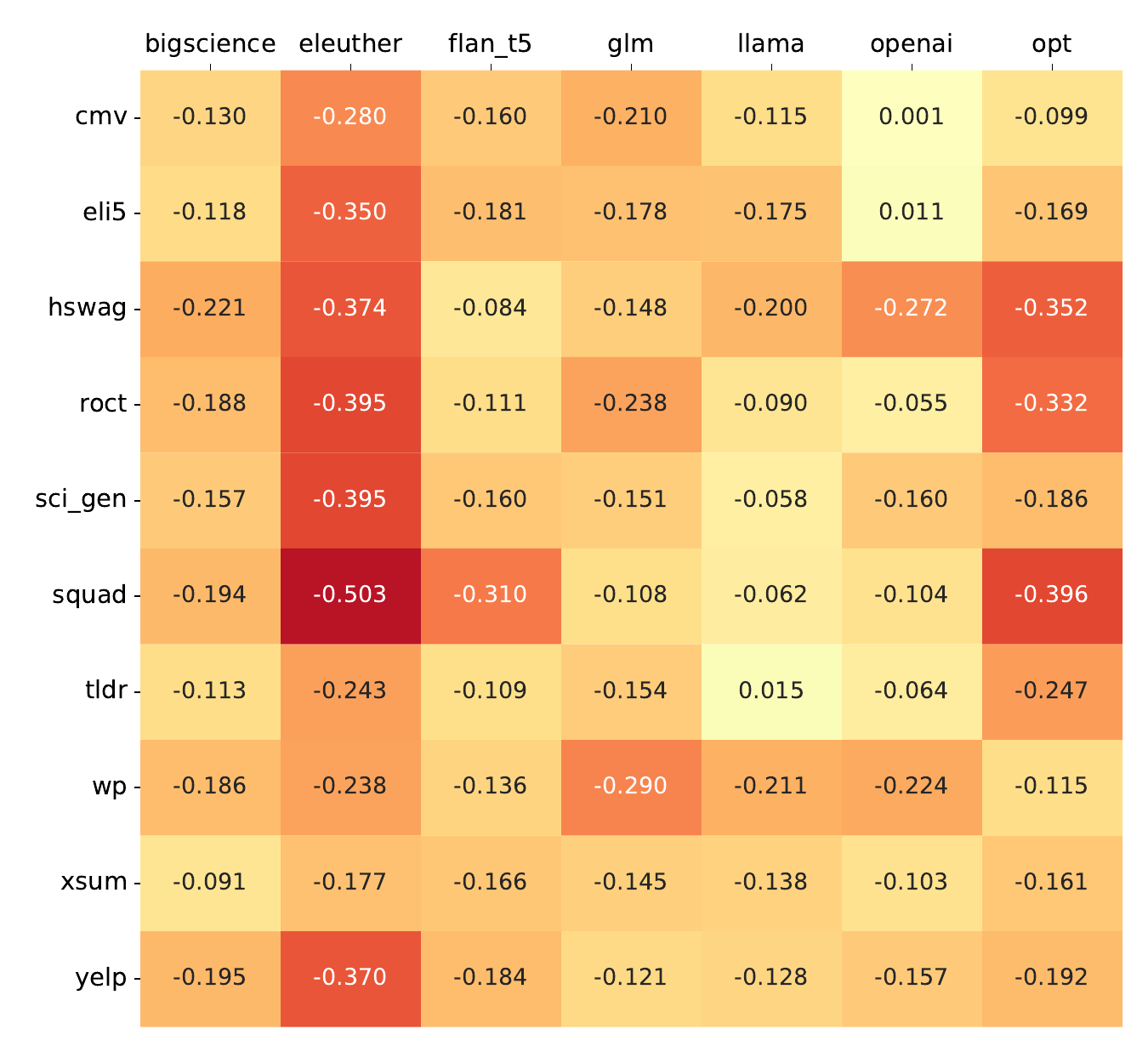}
    \caption{Results of \textbf{TB1.1} domain-model pairs. Positive values (green) indicate lexical richness alone outperforms all 284 features; negative values (red) indicate the full feature set is superior.}
    \label{fig:lexical_only_results_testbed11}
  \end{subfigure}
  \hfill
  \begin{subfigure}[t]{0.48\textwidth}\vspace{0pt}
    \centering
    \includegraphics[width=\textwidth]{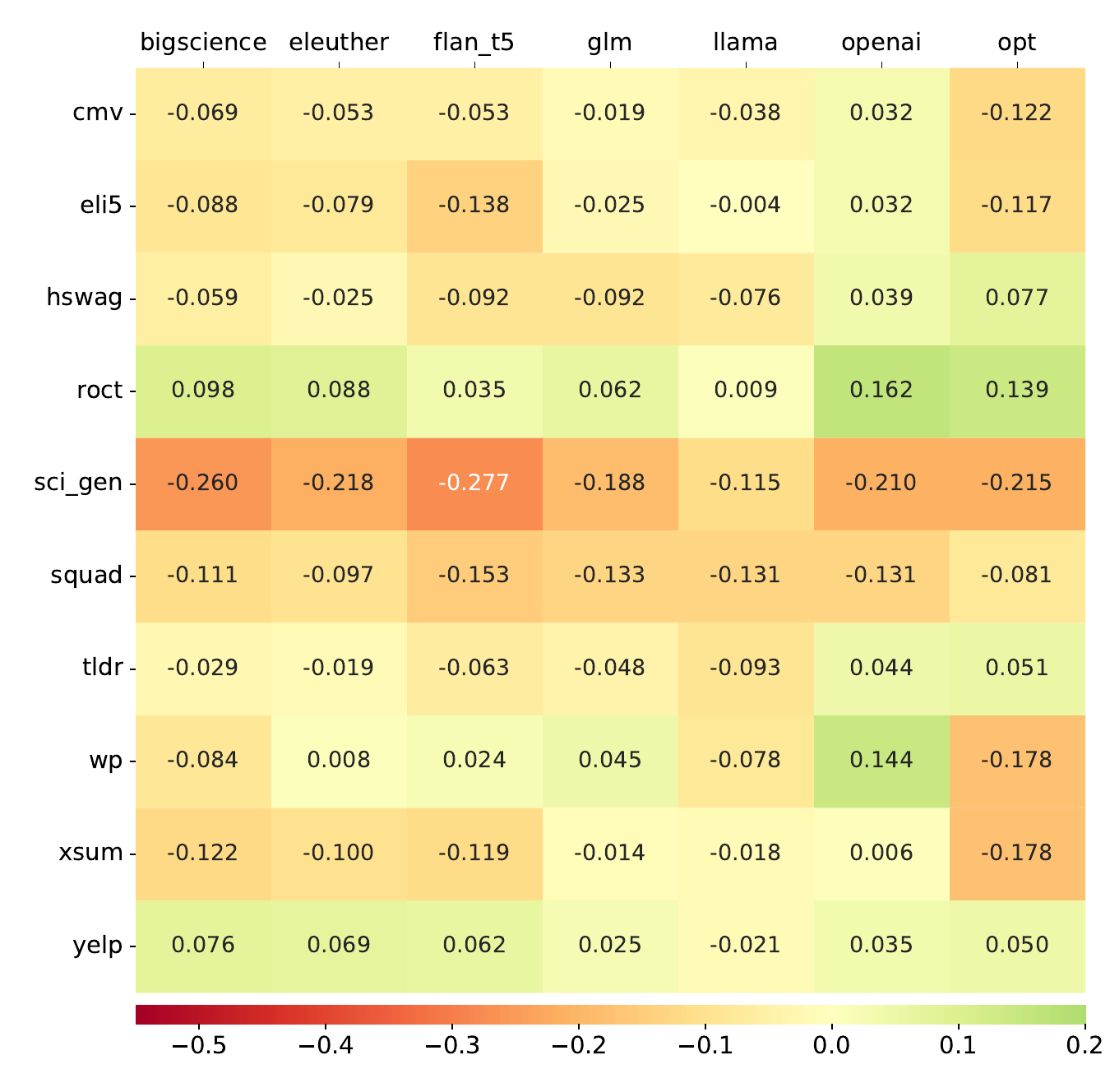}
    \caption{Results of \textbf{TB8} domain-model pairs. Positive values (green) indicate lexical richness alone outperforms all 284 features; negative values (red) indicate the full feature set is superior.}
    \label{fig:lexical_only_results_testbed8}
  \end{subfigure}
  \caption{Performance difference (F1-Macro) between lexical-richness-only and full-feature classifiers.}
  \label{fig:lexical_only_results}
\end{figure*}

We train classifiers using only the three lexical richness features for all 70 text domain--model domain pairs in TB1.1 (ID) and TB8 (OOD), and compare them to the corresponding full-feature baselines. \Cref{fig:lexical_only_results} reports the resulting differences in Macro F1.

\begin{figure}[t]
    \centering
    \includegraphics[width=1\linewidth]{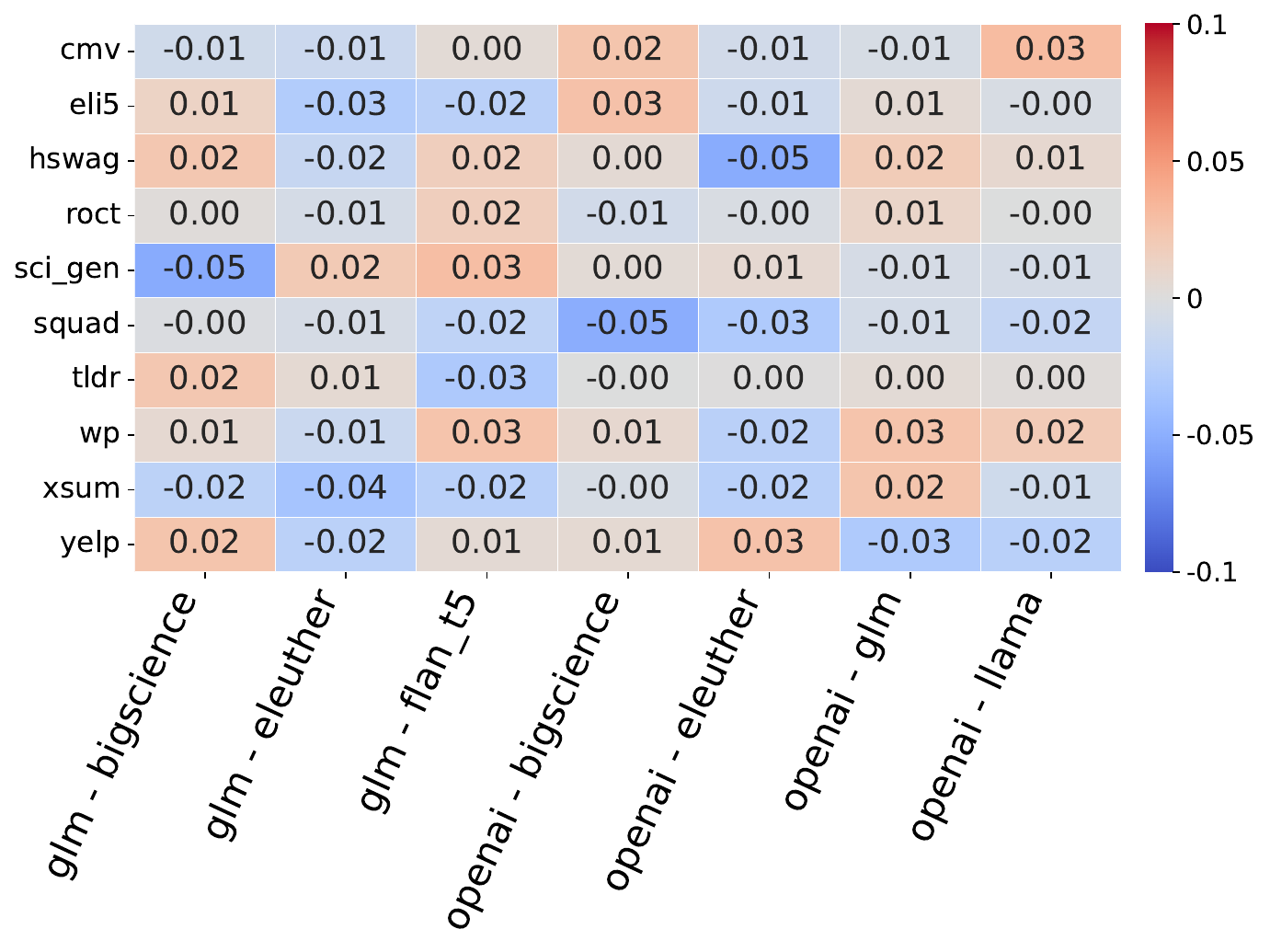}

    \caption{
    Heatmap of pairwise Pearson's $r$ of TTR distributions between selected model family pairs across text domains. The left columns show GLM paired with smaller models, while the right columns show OpenAI paired with models of varying scale. Cell values indicate the Pearson's $r$ between the TTR distributions of the two model families within each text domain.
    }

    \label{fig:pearson_heatmap_ttr_selected_pairs}
\end{figure}

\begin{figure*}[t]
    \centering
    \includegraphics[width=\textwidth]{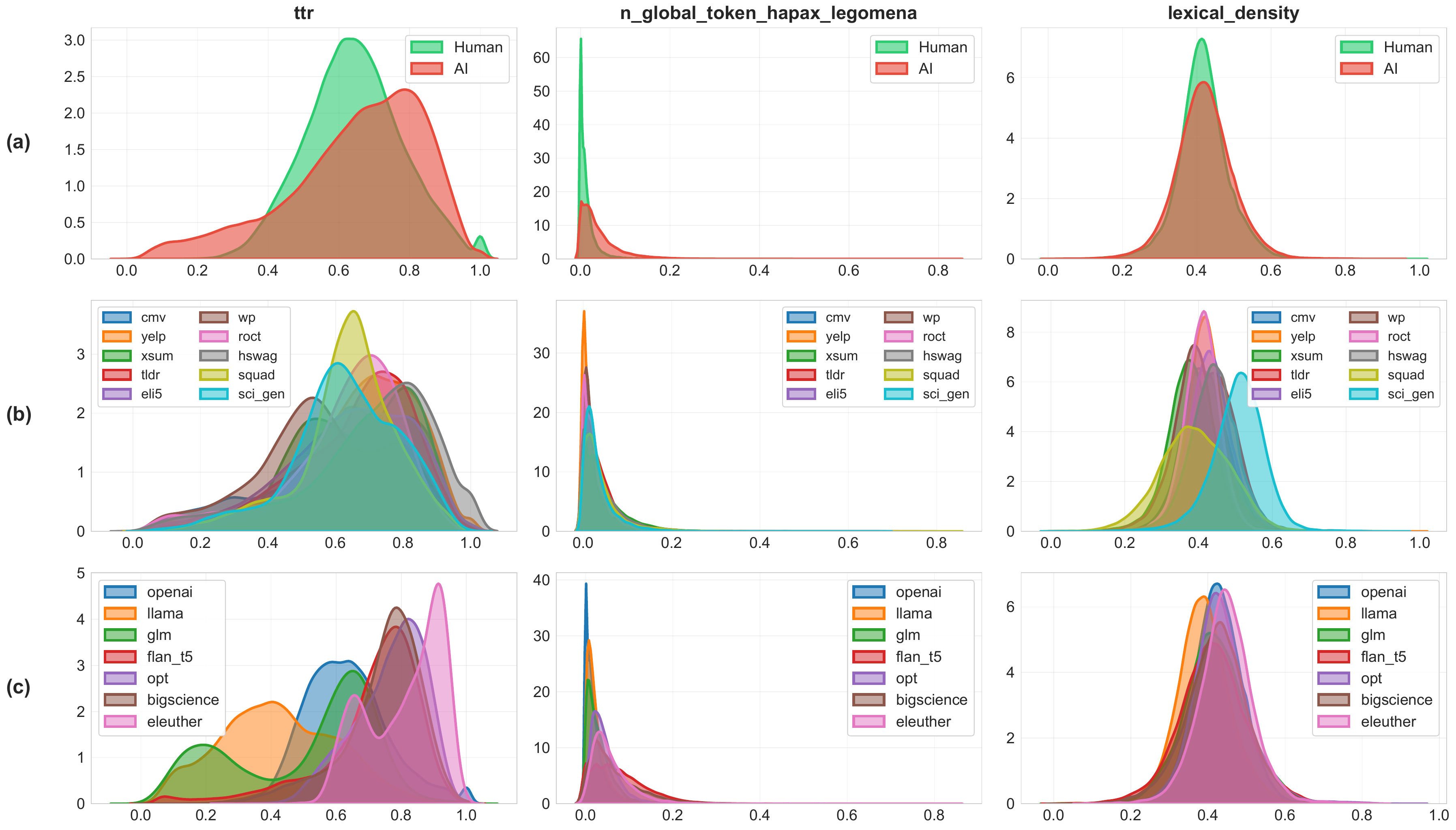}
    \caption{Density distributions of lexical richness features (TTR, hapax legomena, lexical density) across (a) Human vs AI labels, (b) text domains, and (c) AI model domains. Overlapping histograms show the discriminative power of these features across different groupings.}
    \label{fig:feats_distri_analysis_lexical_rich_hist}
\end{figure*}

\begin{figure*}[t]
    \centering
    \includegraphics[width=\textwidth]{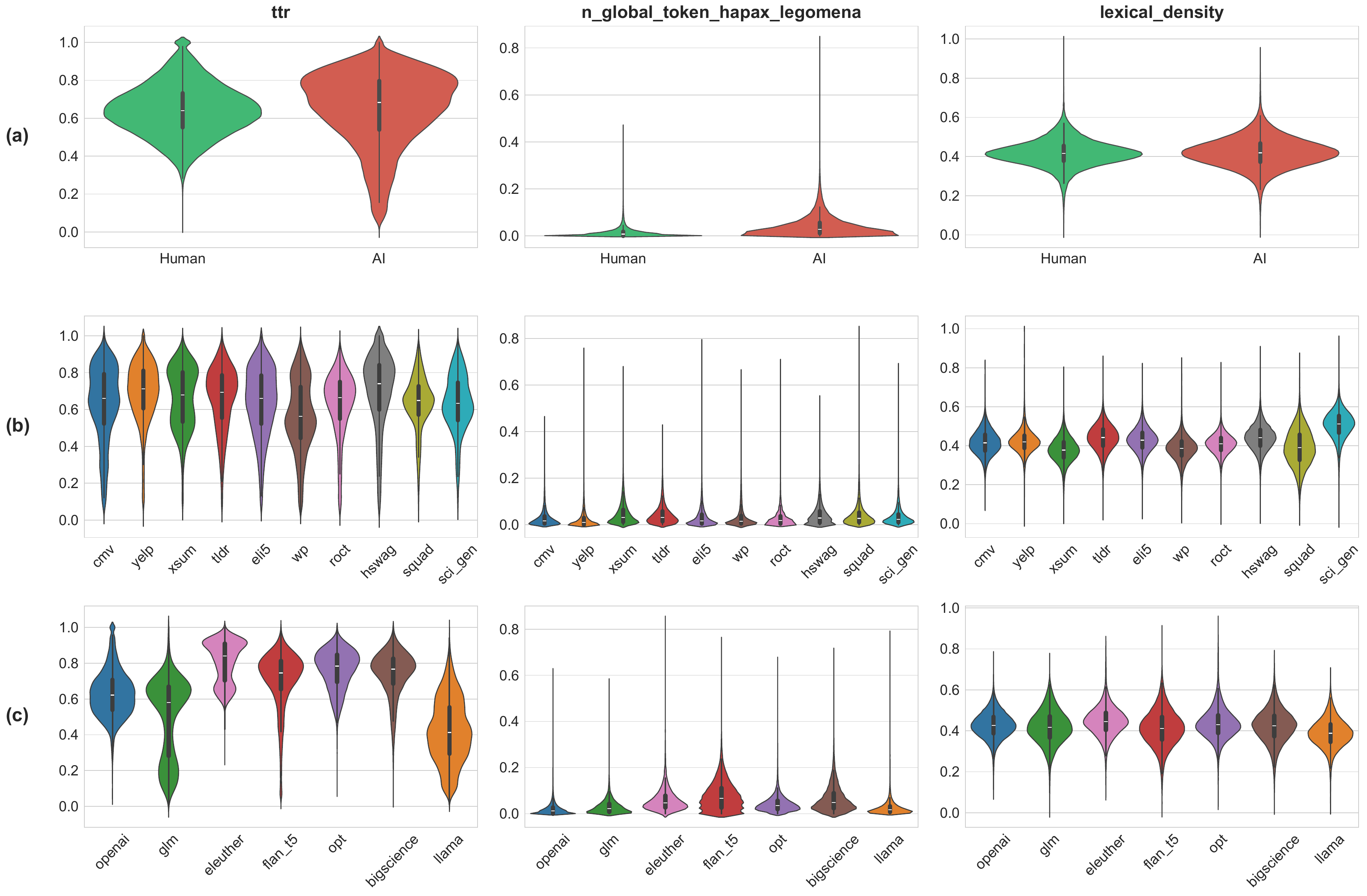}
    \caption{Violin plots for the three lexical richness features: TTR (left), hapax legomena (middle), and lexical density (right). Each row displays distributions across Human vs AI, text domains, and model domains, revealing both central tendency and distributional shape.}
    \label{fig:feats_distri_analysis_lexical_rich_violin}
\end{figure*}

\subsubsection{Feature Distributions Analysis}
\label{sec:feats_distribution_analysis}

Given the substantial impact of lexical richness, we analyze the distributions of its three features: TTR ($n\_types / n\_tokens$), the number of global token hapax legomena (tokens of a text that occur only once in the respective data split; see \Cref{sec:feature-overview}), and lexical density ($n\_lexical\_tokens / n\_tokens$).

\Cref{fig:feats_distri_analysis_lexical_rich_violin,fig:feats_distri_analysis_lexical_rich_hist} show the distributions of the three features by label (row a), text domain (row b), and model domain (row c).

%% file: sections/appendix_qualitative_examples.tex
\begin{figure*}[t]
    \centering
    \begin{subfigure}{1\linewidth}
    \centering
    \includegraphics[height=1.8in]{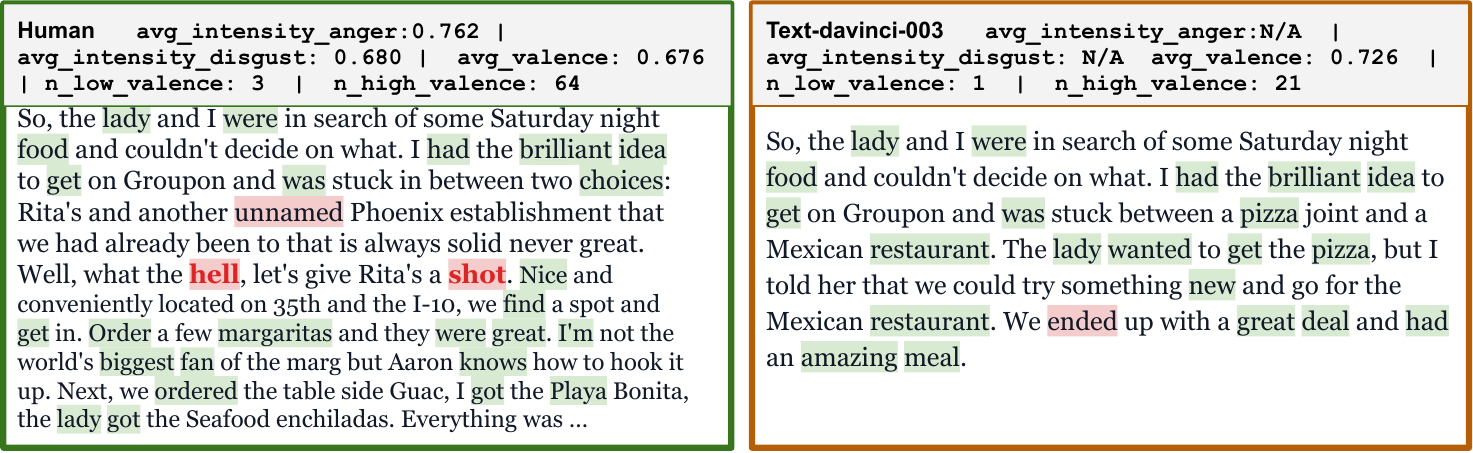}
    \caption{}
    \end{subfigure}%
    \vspace{0.1cm}
    \begin{subfigure}{1\linewidth}
    \centering
    \includegraphics[height=1.8in]{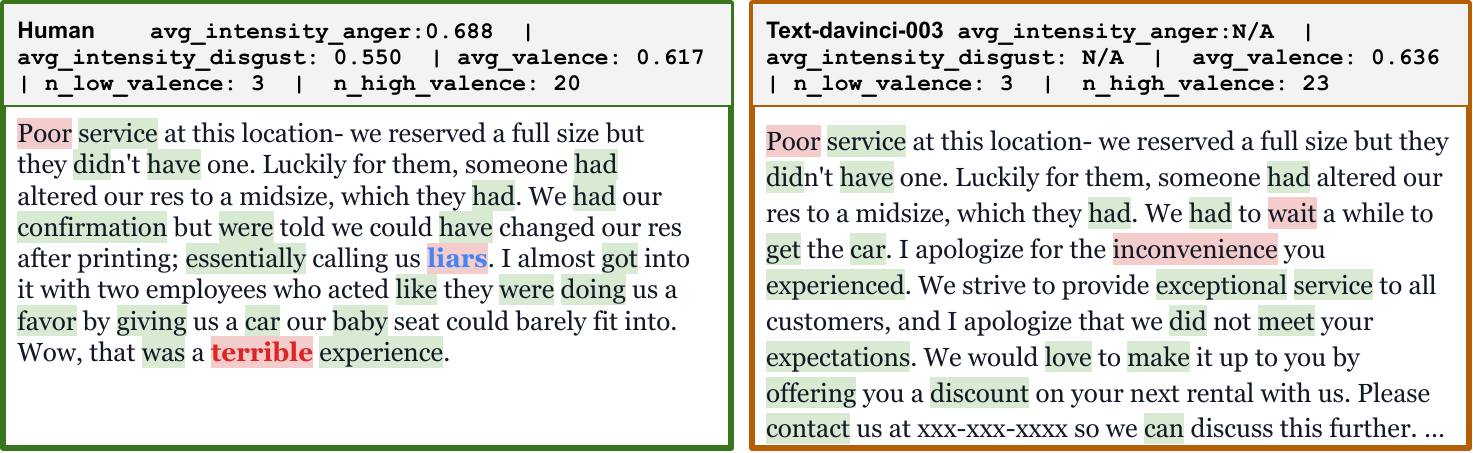}
    \caption{}
    \end{subfigure}%
    \vspace{0.1cm}
    \begin{subfigure}{1\linewidth}
    \centering
    \includegraphics[height=1.8in]{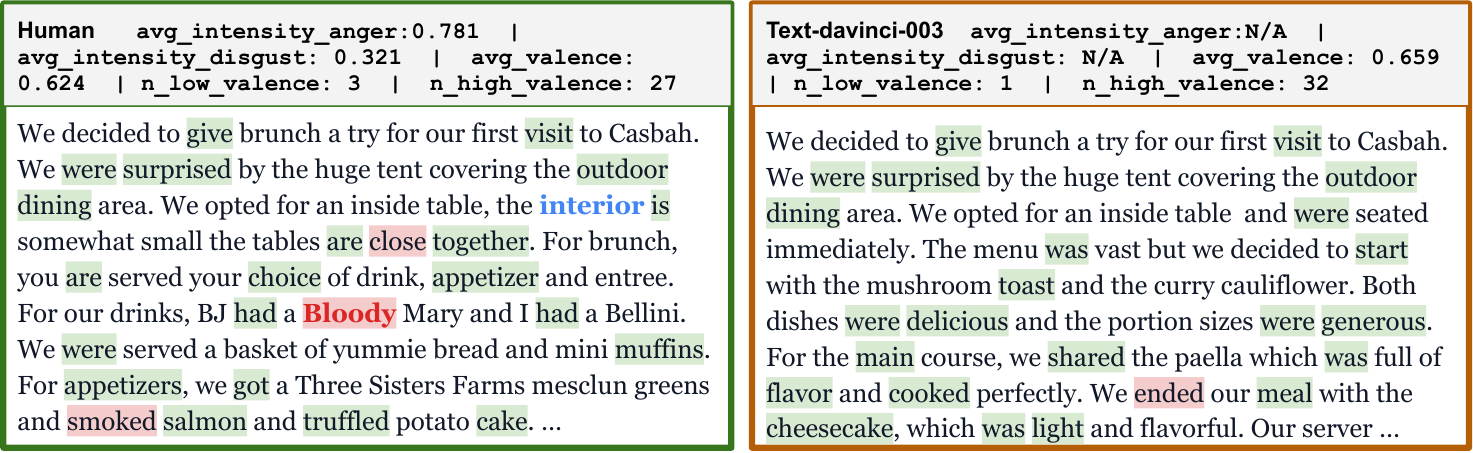}
    \caption{}
    \end{subfigure}
    \caption{Parallel human--AI text pairs from the Yelp domain (AI model: Text-davinci-003) illustrating differences in a few selected features from emotion area. Visual markers: \textbf{{\color{red} red bold text}} = anger intensity, \textbf{{\color{disgustblue} blue bold text}} = disgust intensity, \colorbox{highlightgreen}{green highlight} = high valence ($>0.66$) and \colorbox{red!20}{pink highlight} = low valence ($<0.33$). feature values (average anger and disgust  intensity, average valence, and high/low valence word counts) are shown above each text. Emotion intensity is derived from the NRC Affect Intensity Lexicon and valence from the NRC VAD lexicon, following the \texttt{elfen} toolkit's extraction procedure. Some texts are truncated for space; ellipses (\dots) indicate omitted content.}
    \label{fig:example-pairs-emotion}
\end{figure*}

\begin{figure*}[t]
    \centering
    \begin{subfigure}{1\linewidth}
    \centering
    \includegraphics[height=1.8in]{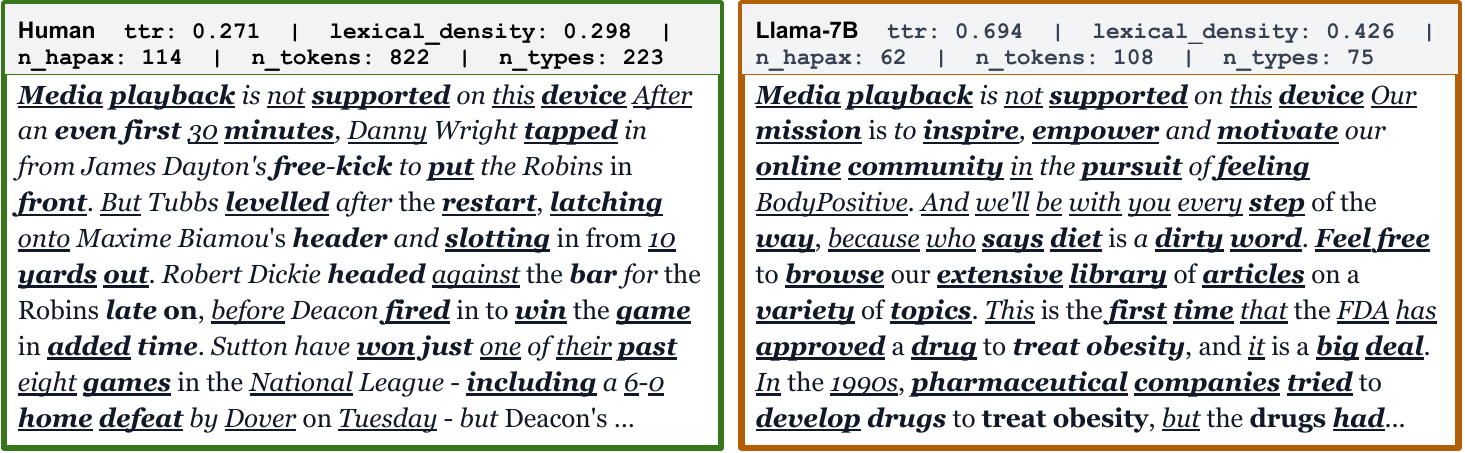}
    \caption{Human (Left) vs. LLaMA-7B (right).}
    \end{subfigure}%
    \vspace{0.1cm}
    \begin{subfigure}{1\linewidth}
    \centering
    \includegraphics[height=1.8in]{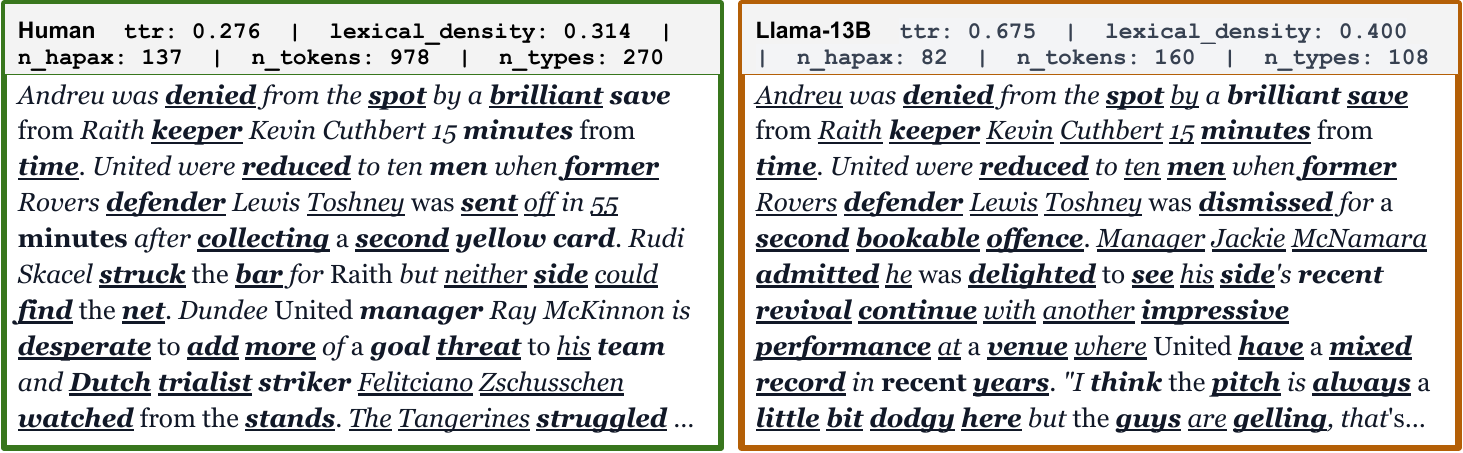}
    \caption{Human (left) vs. LLaMA-13B (right).}
    \end{subfigure}%
    \vspace{0.1cm}
    \begin{subfigure}{1\linewidth}
    \centering
    \includegraphics[height=1.8in]{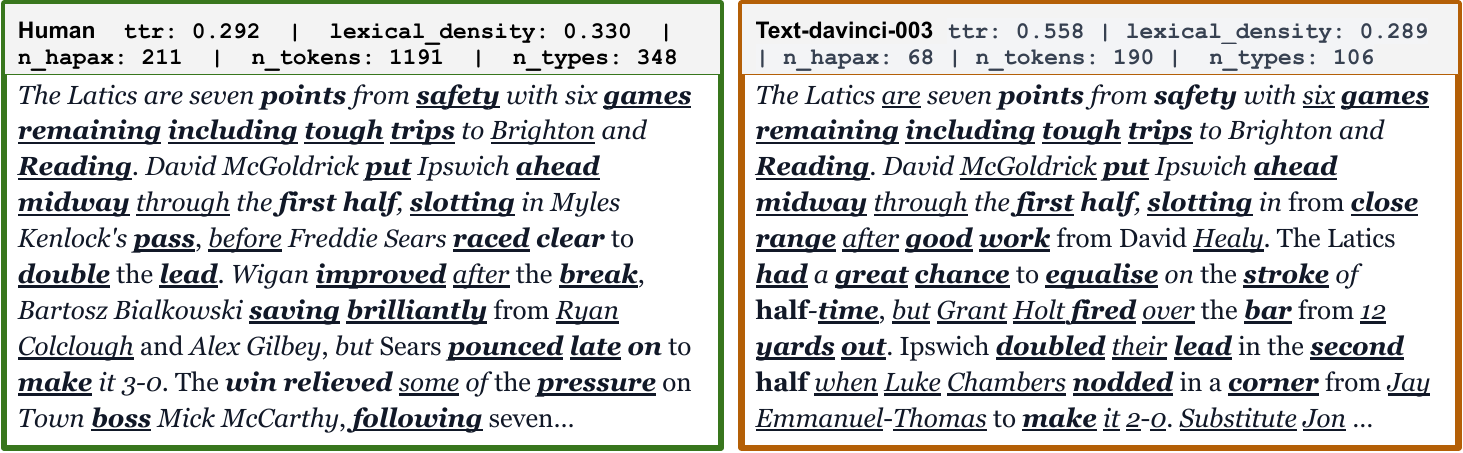}
    \caption{Human (left) vs. Text-davinci-OO3 (right).}
    \end{subfigure}
    \caption{Parallel human-AI text pairs from the XSum domain illustrating differences in lexical richness features. Each pair shares the same opening words, with the AI text generated as a continuation. Visual markers: \textit{italic}= first occurrence of a token (unique type, contributing to TTR), \textbf{bold} = lexical token (noun, verb, adjective, or adverb, contributing to lexical density, and \underline{underline} = hapax legomenon (token appearing only once in the text). Feature values (TTR, lexical density, hapax count, token count, and type count) are shown above each text. Texts are truncated for space; ellipses (\dots) indicate omitted content. }
    \label{fig:example-pairs-lexical-richness}
\end{figure*}

To complement our quantitative analysis, we present qualitative examples that illustrate how the extracted linguistic features manifest in human-written versus AI-generated text. We select two feature areas--Emotion and Lexical Richness-- based on their performance and clear distinction in both texts, and choose the domain-model pairs that best highlight the contrast identified in our experiments.

\subsection{Emotion Features --- Domain: Yelp, Model: OpenAI text-davinci-003}
We select the Yelp domain paired with OpenAI's \texttt{text-davinci-003} for the emotion feature visualization (\Cref{fig:example-pairs-emotion}), as Yelp reviews are inherently opinion-driven and exhibit a wide range of emotional expression among the text domains. For each sample, we display two emotion intensity dimensions—\texttt{anger} ({\color{red!60}red}) and \texttt{disgust} ({\color{disgustblue}blue})—where darker shading indicates higher intensity scores from the NRC Affect Intensity Lexicon \citep{mohammad-2018-word}. Valence is encoded background highlighting: \colorbox{highlightgreen}{green} indicates high valence ($> 0.66$) and \colorbox{red!20}{pink} indicates low valence ($< 0.33$) based on the NRC VAD Lexicon \citep{mohammad-2018-obtaining}. Aggregate statistics (\texttt{avg\_intensity}, \texttt{avg\_valence}, \texttt{n\_low/high\_valence}) are shown above each sample.

\subsection{Lexical Richness --- Domain: XSum, Models: LLaMA \& OpenAI}

We select XSum paired with LLaMA (-7B,-13B) and OpenAI (\texttt{text-davinci-003}) for the lexical richness visualization (\Cref{fig:example-pairs-lexical-richness}), as these pairs (combinations of domain and models) exhibited one of the largest performance drops in TB8 when lexical richness features were removed (cf.\ \Cref{fig:detailed_results_of_ablation_for_16_domain_domain_pair}), and the spider plot analysis (\Cref{fig:spiderweb_plot_ablation_for_16_domain_domain_pairs_app}) confirmed that lexical richness is the dominant feature area for these pairs. Furthermore, XSum (news summarization) produces structurally diverse texts where lexical patterns differ markedly between human and AI writing.

Each token is annotated along three dimensions: \textit{italic} marks first occurrences (unique types contributing to TTR), \textbf{bold} marks lexical tokens (nouns, verbs, adjectives, and adverbs contributing to lexical density), and \underline{underline} marks hapax legomena (tokens appearing only once in the text). Aggregate statistics (\texttt{ttr}, \texttt{lexical\_density}, \texttt{n\_hapax\_legomena}, \texttt{n\_tokens}, \texttt{n\_types}) are displayed above each sample.

%% file: sections/appendix_D.tex
\begin{figure*}[t]
    \centering
    \includegraphics[width=\textwidth]{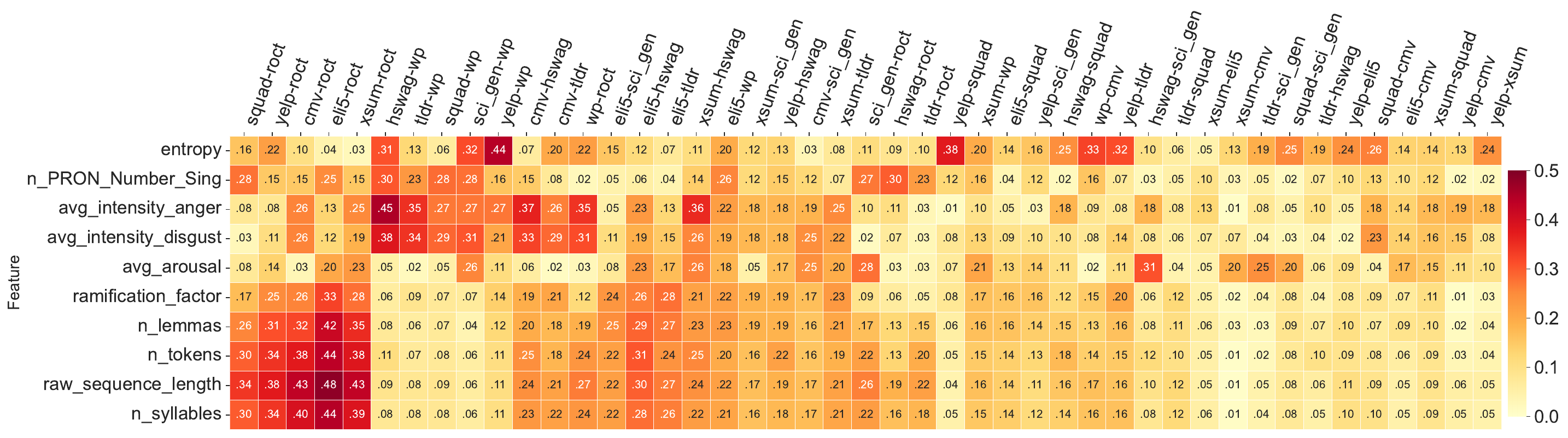}

    \caption{Clustered heatmap of pairwise Wasserstein distances for the top 10 most discriminative linguistic features across all human text domain pairs. Rows represent features that most consistently appear in the top 10 most discriminative features across domain pairs, ranked by their frequency of occurrence. Columns represent each pair of domains. Higher values (darker cells) indicate greater distributional divergence between the two domains for that feature, suggesting that the feature captures domain-specific linguistic patterns. Lower values (lighter cells) indicate similar distributions across domains, suggesting domain-invariant characteristics}
    \label{fig:wasserstein_common_top10_heatmap_text_domains}
\end{figure*}

\subsection{Wasserstein Distance of Text Domains}
\label{sec:wasserstein_text_domains_appendix}
\Cref{fig:wasserstein_common_top10_heatmap_text_domains} reports pairwise Wasserstein distances between human-written text domains for the ten most discriminative features. The domain pairs form two groups. Pairs involving ROCStories, HellaSwag, WritingPrompts, and TLDR show higher divergence, mainly in surface-level features (e.g., \texttt{raw\_sequence\_length}, \texttt{n\_tokens}, \texttt{n\_syllables}, \texttt{n\_lemmas}); three of these domains show the largest performance drops in TB6 (\Cref{tab:baseline_testbed3_vs_6_comparison}). Pairs involving CMV, XSum, ELI5, SQuAD, and Yelp show lower divergence, in line with their smaller OOD drops.

\texttt{entropy} and \texttt{n\_PRON\_Number\_Sing} follow a different pattern, with higher divergence for pairs involving Yelp and SQuAD. This suggests that, while surface features drive most between-domain differences, other feature areas capture complementary domain-specific signals.

\begin{figure*}[t]
    \centering
    \includegraphics[width=\textwidth]{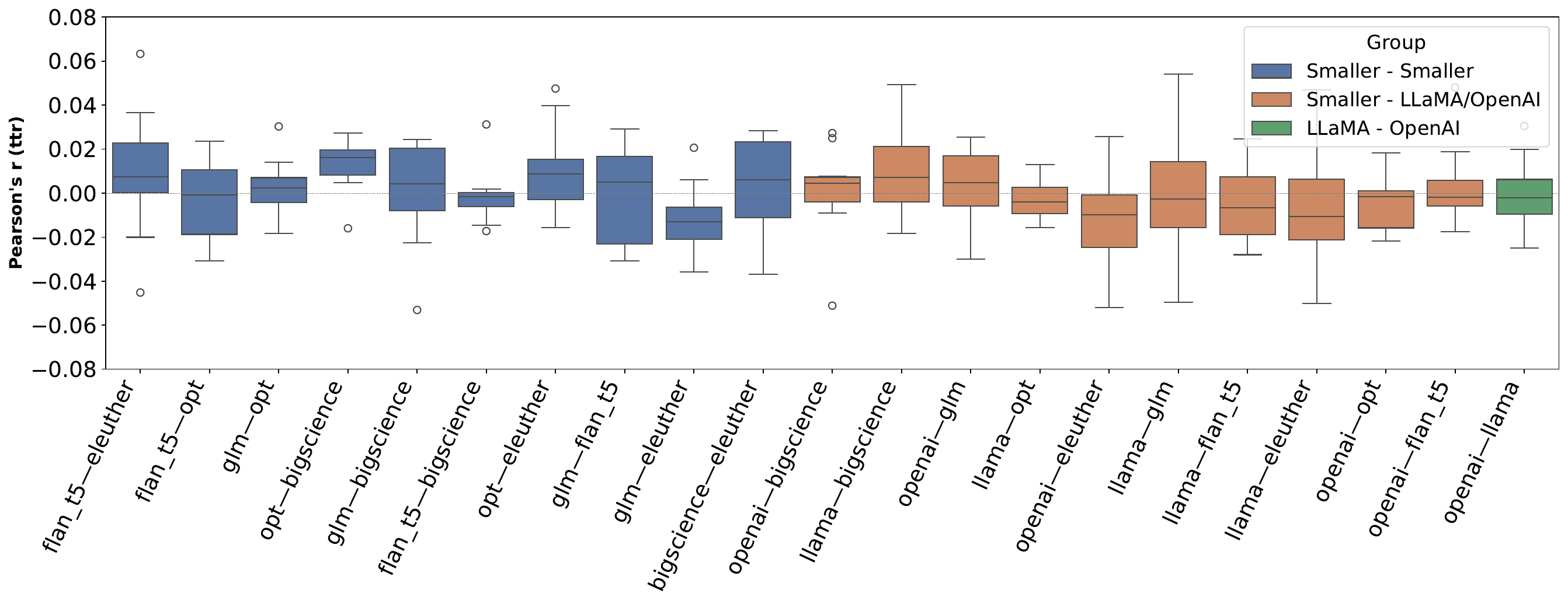}

    \caption{Pairwise Pearson's $r$ of TTR feature distributions across model family pairs, with distributions computed over the ten text domains. Pairs are grouped into three categories: \textit{Smaller--Smaller} (blue; e.g., FLAN-T5 vs. OPT), \textit{Smaller--LLaMA/OpenAI} (orange; e.g., Eleuther vs. LLaMA), and \textit{LLaMA--OpenAI} (green). TTR was selected as the focal feature given its pronounced human--AI distributional separation observed in \Cref{sec:feats_distribution_analysis}.
    }
    \label{fig:pearson_ttr_family_by_comparison}
\end{figure*}

\subsection{Pearson's $r$ Calculations of Model Domains}
\label{sec:pearsonr_model_families_appendix}

Building on the distributional analysis in \Cref{sec:feats_distribution_analysis}, we compute pairwise Pearson's $r$ between the TTR distributions of each pair of model families within each text domain, yielding ten $r$ values per pair. We group the pairs into three categories following the ablation results (\Cref{results:ablation}): \textit{Smaller--Smaller} (e.g., FLAN-T5 vs.\ OPT), \textit{Smaller--LLaMA/OpenAI} (e.g., EleutherAI vs.\ LLaMA), and \textit{LLaMA--OpenAI}.

\Cref{fig:pearson_ttr_family_by_comparison} reports the distributions of $r$ for all 21 model family pairs. The correlations are close to zero for all pairs, with no clear differences between the three groups, indicating that the TTR patterns of model families do not co-vary across text domains. This is consistent with the model-dependent effects of lexical richness in the ablation studies (\Cref{results:ablation}).

\Cref{fig:pearson_heatmap_ttr_selected_pairs} reports the per-domain values for selected pairs that reflect contrasts identified in the ablation analysis: GLM paired with smaller models (left) and OpenAI paired with models of different scale (right). Correlations remain close to zero across all text domains, without systematic domain-specific deviations.